\documentclass[lettersize,journal]{IEEEtran}
\usepackage{amsmath,amsfonts}
\usepackage{algorithmic}
\usepackage{algorithm}
\usepackage{array}
\usepackage{textcomp}
\usepackage{stfloats}
\usepackage{url}
\usepackage{verbatim}
\usepackage{graphicx}
\usepackage{cite}
\usepackage{multirow}
\usepackage{rotating}
\usepackage{makecell,diagbox}
\usepackage{float}
\usepackage{subfigure}
\usepackage{bm}
\usepackage{booktabs}
\usepackage{hyperref}
\usepackage{xcolor}

\begin{document}

\title{Prompt-Driven Lightweight Foundation Model for Instance Segmentation-Based Fault Detection in Freight Trains}

\author{
	\vskip 1em
	
	Guodong Sun, Qihang Liang, Xingyu Pan, Moyun Liu and Yang Zhang

	\thanks{
	
        (Corresponding author: Yang~Zhang)
		
		Guodong Sun, Qihang Liang, Xingyu Pan, and Yang Zhang are with the School of Mechanical Engineering, Hubei University of Technology, Wuhan 430068, China. And the Hubei Key Laboratory of Modern Manufacturing Quality Engineering, Hubei University of Technology, Wuhan 430068, China.(e-mail: sunguodong@hbut.edu.cn; liangqihang@hbut.edu.cn; 502200007@hbut.edu.cn; yzhangcst@hbut.edu.cn)
		
		Moyun Liu is with the School of Mechanical Science and Engineering, Huazhong University of Science and Technology, Wuhan 430074, China.(e-mail: lmomoy@hust.edu.cn)
	}
}
\markboth{Journal of \LaTeX\ Class Files,~Vol.~14, No.~8, August~2021}%
{Shell \MakeLowercase{\textit{et al.}}: A Sample Article Using IEEEtran.cls for IEEE Journals}

\maketitle

\begin{abstract}
Accurate visual fault detection in freight trains remains a critical challenge for intelligent transportation system maintenance, due to complex operational environments, structurally repetitive components, and frequent occlusions or contaminations in safety-critical regions. Conventional instance segmentation methods based on convolutional neural networks and Transformers often suffer from poor generalization and limited boundary accuracy under such conditions.
To address these challenges, we propose a lightweight self-prompted instance segmentation framework tailored for freight train fault detection. Our method leverages the Segment Anything Model by introducing a self-prompt generation module that automatically produces task-specific prompts, enabling effective knowledge transfer from foundation models to domain-specific inspection tasks. In addition, we adopt a Tiny Vision Transformer backbone to reduce computational cost, making the framework suitable for real-time deployment on edge devices in railway monitoring systems.
We construct a domain-specific dataset collected from real-world freight inspection stations and conduct extensive evaluations. Experimental results show that our method achieves 74.6 $AP^{\text{box}}$ and 74.2 $AP^{\text{mask}}$ on the dataset, outperforming existing state-of-the-art methods in both accuracy and robustness while maintaining low computational overhead.
This work offers a deployable and efficient vision solution for automated freight train inspection, demonstrating the potential of foundation model adaptation in industrial-scale fault diagnosis scenarios. Project page: \url{https://github.com/MVME-HBUT/SAM_FTI-FDet.git}
\end{abstract}

\section{Introduction}
\label{sec:introduction}
\IEEEPARstart{I}{n} modern freight train inspection systems, ensuring reliable and efficient fault detection is critical for maintaining operational safety and minimizing economic losses. Key components such as brake shoes and bearing saddles are particularly prone to excessive wear, yet traditional object detection methods lack the capability to quantitatively analyze these wear conditions \cite{wei2019defect}. When such faults remain undetected, they can result in severe safety incidents or costly service disruptions. While recent studies have also explored signal-based monitoring techniques \cite{rahman2024remote} and deep learning–based defect classification for railway maintenance \cite{lu2022rail}, visual inspection remains the most direct and informative approach for assessing mechanical conditions. However, these systems must still contend with diverse fault types, complex mechanical structures, and harsh outdoor environments, which present substantial challenges for conventional manual inspection. Although deep learning–based visual fault detection methods have been widely deployed in freight train fault detection systems \cite{zhang2021unified}, they still face critical challenges in practical applications, including insufficient accuracy, limited generalization capability, and poor adaptability for detecting specialized components. 

\begin{figure}[!t]
    \centering
    \subfigure[]{
    \includegraphics[width=0.24\columnwidth]{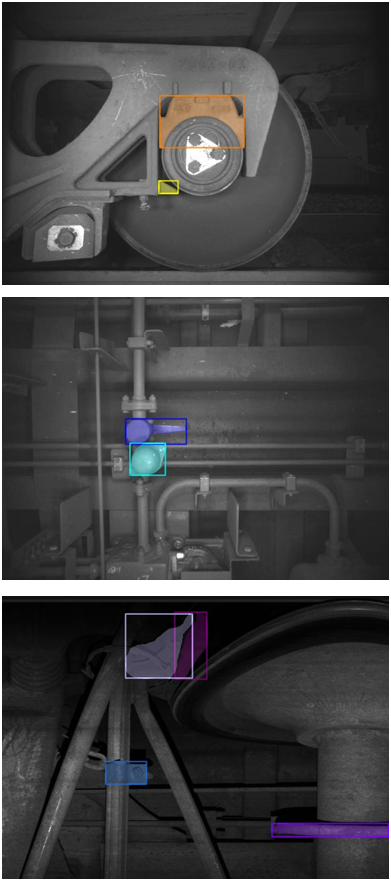}  
    } \hspace{-1.3em}
    \subfigure[]{
    \includegraphics[width=0.24\columnwidth]{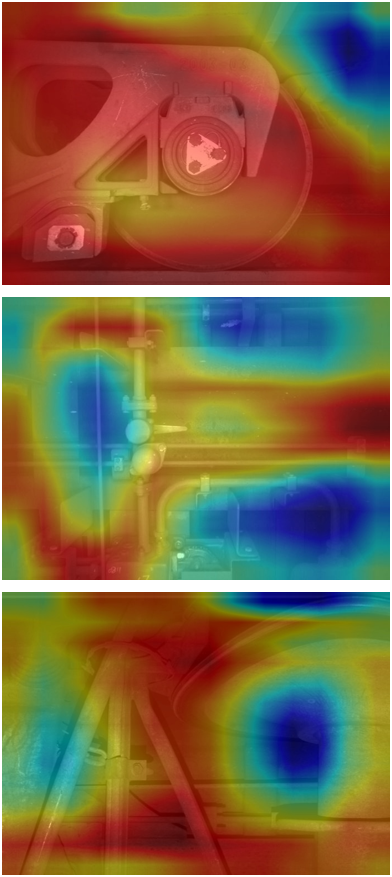}  
    } \hspace{-1.3em}
    \subfigure[]{
    \includegraphics[width=0.24\columnwidth]{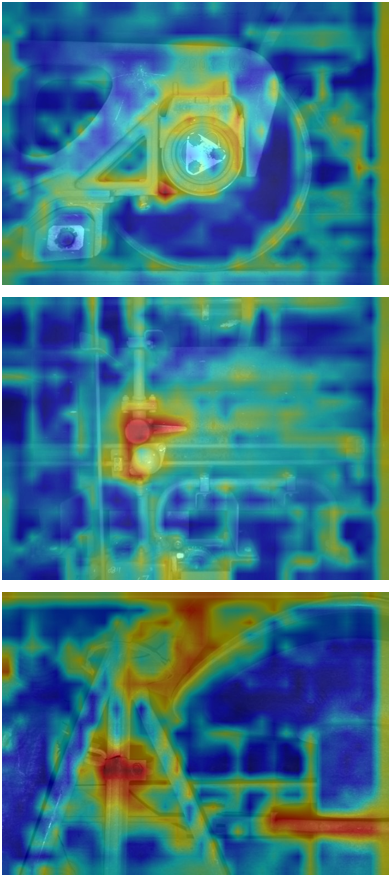}  
    } \hspace{-1.3em}
    \subfigure[]{
    \includegraphics[width=0.24\columnwidth]{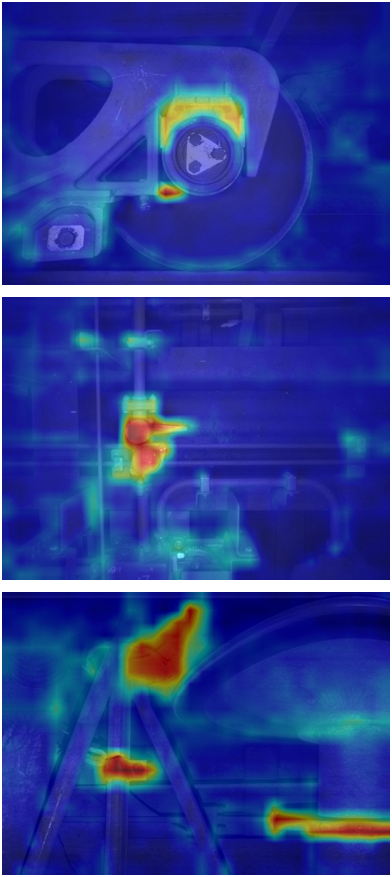}  
    }
        \caption{Visualization results of activation maps from various detectors. (a) Ground truth annotations for each image. (b) Activation maps of Mask R-CNN. (c) Activation maps of SAM. (d) Our proposed prompter-driven SAM detection method. As the activation maps exemplify, our method can more accurately capture the region of interest due to the efficient prompt mechanism.}
	\label{heatmap}
\end{figure}

However, despite these advantages, existing deep learning approaches still face significant industrial challenges in terms of scalability, generalization, and deployment efficiency. For instance, Zhang et al. \cite{zhang2022visual} introduced a lightweight anchor-free detection framework tailored to key braking components. While the method delivers high accuracy with minimal computational cost, the authors also reported substantial performance degradation when the model was applied to previously unseen inspection stations. This clearly illustrates that its scenario-dependent design constrains its robustness and limits its ability to generalize across different domains. Similarly, Feng et al. \cite{feng2024integrating} observed misidentification issues under out-of-distribution conditions, further indicating that conventional CNN-based detection frameworks are sensitive to domain shifts.
Similarly, recent approaches have explored spatial-aware dynamic distillation techniques that incorporate multilayer perceptrons and axial-shift mechanisms to capture semantic information without relying on pre-trained teachers \cite{zhang2024spatial}. Although these strategies enhance feature modeling and detection adaptability, they often rely on complex training pipelines, show sensitivity to label quality, and impose high computational overhead, which makes them difficult to scale in real-time, resource-constrained industrial systems. 
Despite their technical strengths, these methods are typically developed on curated datasets with limited diversity, focusing on isolated fault categories. As a result, they struggle to generalize to rare or evolving fault types in dynamic railway environments \cite{feng2024integrating}. This lack of robustness and generalization remains a key bottleneck preventing their widespread adoption in industrial freight inspection applications.

At the same time, the emergence of foundation models, which are large-scale pretrained models with strong generalization and transfer learning capabilities, has transformed the field of visual understanding \cite{chen2024rsprompter}. One prominent example is the Segment Anything Model (SAM) \cite{kirillov2023segment}, trained on over one billion segmentation masks and capable of performing class-agnostic segmentation using various types of input prompts. SAM demonstrates excellent performance across a wide range of segmentation tasks. Nonetheless, its dependence on external prompts (e.g., clicks or bounding boxes) and sensitivity to prompt placement make it unsuitable for fully automated industrial applications. When applied directly to freight train images, the segmentation capability of SAM is further constrained by the inherent structural complexity, domain shift, and ambiguous contours.

To address these limitations and effectively transfer the knowledge embedded in foundation models to the domain of freight train fault detection, we propose a SAM-based instance segmentation framework enhanced with a novel automatic prompting strategy (SAM FTI-FDet). Central to our method is the Prompt generator, which eliminates the need for manual intervention by learning to generate high-quality prompts directly from global image representations. These prompts enable SAM to produce class-relevant instance masks in a fully automated manner, tailored to the structural semantics of freight train components.

Furthermore, we incorporate the lightweight TinyViT-SAM backbone from MobileSAM \cite{zhang2023faster} to significantly reduce computational cost and memory usage, enabling deployment on resource-constrained edge devices typically found in real-world railway monitoring systems. To further facilitate efficient knowledge transfer, we introduce an end-to-end set-based prediction mechanism, wherein a set of fixed and learnable queries directly extract task-relevant spatial and semantic information. As shown in Fig. \ref{heatmap} , this design not only bridges the gap between general foundation knowledge and domain-specific tasks but also enhances the robustness and precision of instance segmentation under challenging visual conditions.

Our contributions can be summarized as follows:
\begin{enumerate}
    \item We explored the use of self-prompt methods to explicitly express domain knowledge and prior information, transferring the foundation model's pre-trained large-scale knowledge base to specific domains.
    \item We designed an efficient SAM-based approach suitable for resource-constrained scenarios, while achieving knowledge transfer and precise domain knowledge injection by integrating globally pre-trained general knowledge and separating local domain-specific features.
    \item We proposed a novel prompt generation method based on an end-to-end set prediction mechanism, which generates a set of fixed and learnable query prompts. These prompts enable the model to directly extract highly task-relevant information from global image features.
\end{enumerate}

This work demonstrates that by leveraging the knowledge distillation and adaptation capabilities of large-scale foundation models, it is possible to significantly improve the automation, accuracy, and efficiency of fault detection systems in the railway domain, offering a new paradigm for intelligent transportation system maintenance.
\begin{figure*}[!t]
    \centering
    \includegraphics[width=7.1in]{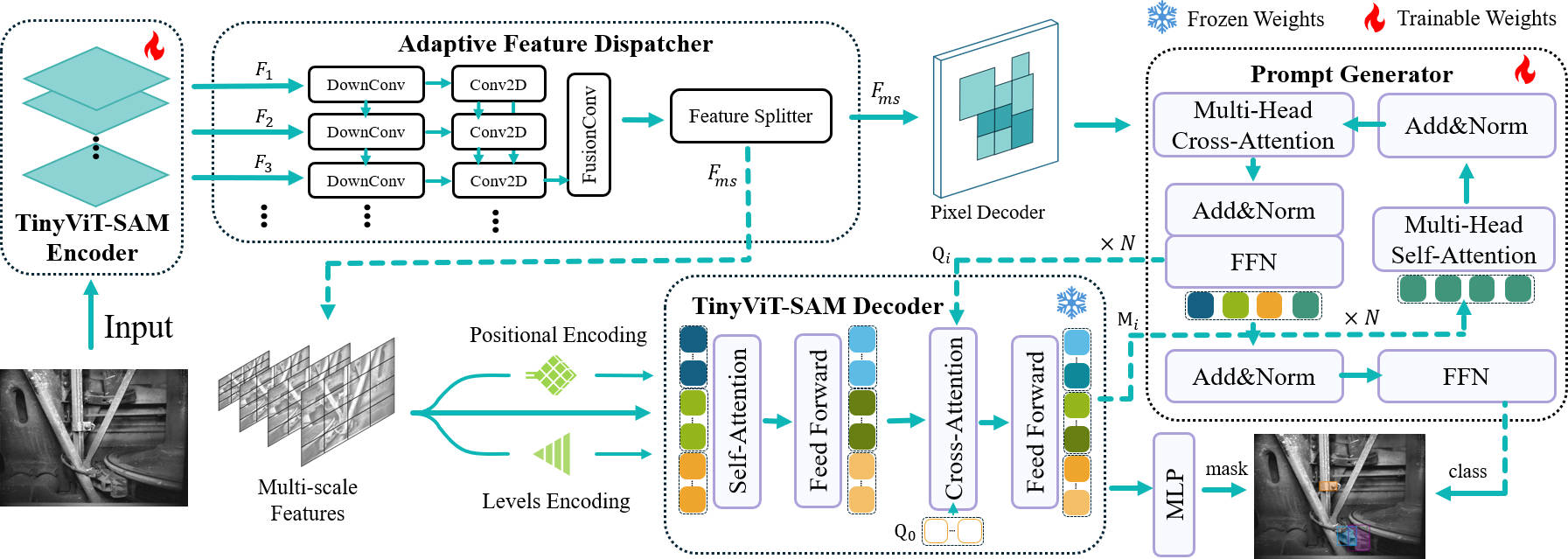}
    \caption{An overview of our proposed SAM-based framework for visual fault detection of freight trains is presented. The network is designed to leverage a prompt generator to provide high-quality prompts for TinyViT-SAM, thereby enhancing the segmentation capability of the model. Furthermore, spatial and hierarchical embeddings are incorporated to improve the image perception ability of TinyViT-SAM, enabling efficient and low-cost optimization.} 
	\label{structure}
\end{figure*}

\section{Related Work}
\subsection{Fault Detection for Freight Train Images}
Freight train fault detection has evolved with computer vision integration, progressing from traditional methods using manually designed features to deep learning approaches that automatically extract high-level features. While traditional methods like Liu et al.'s cascaded detection lacked flexibility \cite{liu2015automated}, deep learning systems such as Chen et al.'s end-to-end framework \cite{chen2020exploring} and Zhou et al.'s NanoDet-based system \cite{ye2021fault} show promise but often target specific components. CNN-based systems \cite{sun2017automatic} have improved fault recognition performance, but typically rely on complex network architectures that are unsuitable for real-time industrial applications.

As deep learning models continue to evolve, more advanced vision tasks have been introduced into fault detection systems, among which instance segmentation has emerged as a promising solution for precise component-level analysis. Instance segmentation combines object detection and semantic segmentation for pixel-level predictions. Two-stage approaches like Mask R-CNN \cite{he2017mask} generate region proposals before mask prediction. Single-stage methods offer improved efficiency, with YOLACT \cite{bolya2019yolact} pioneering real-time solutions and SOLO \cite{wang2020solo}/SOLOv2 \cite{wang2020solov2} reformulating the task as direct mask prediction. Recent innovations include Mask2Former \cite{cheng2022masked}, which unifies various segmentation tasks through transformer decoder architecture; PointRend \cite{kirillov2020pointrend}, which refines mask boundaries using high-resolution rendering; CondInst \cite{tian2020conditional}, which dynamically generates instance masks; SparseInst \cite{Cheng2022SparseInst}, leveraging sparse convolutions to reduce computational complexity; RTMDet \cite{lyu2212rtmdet}, optimizing both detection and segmentation; and QueryInst \cite{Fang_2021_ICCV}, which transforms segmentation into a query selection task.

Foundation models like SAM offer new possibilities but face challenges in industrial applications due to complex environments and specific precision requirements. Object detection approaches struggle with quantitative analysis of defects in components such as bearing saddles and brake elements, highlighting the need for specialized approaches that combine deep learning-based segmentation with freight train fault detection for efficient and adaptable real-time inspection systems.

\subsection{Foundation Model}
Foundation models have revolutionized artificial intelligence with strong generalization through large-scale pretraining. Language models such as ChatGPT \cite{brown2020language} and DeepSeek \cite{wu2024deepseek} show impressive performance in natural language processing, while vision models like DALL-E \cite{ramesh2021zero} and Stable Diffusion \cite{rombach2022high} excel in image generation. Vision-language models like CLIP \cite{radford2021learning} have also set new benchmarks in visual recognition.

The SAM represents a major breakthrough in computer vision by introducing a prompt-based segmentation paradigm. Trained on over one billion masks and eleven million images, SAM demonstrates strong generalization across diverse segmentation tasks \cite{kirillov2023segment}. Its architecture supports multiple input prompts, including points, bounding boxes, and text, making it versatile for a range of applications. To address its high computational demands, lightweight variants have been developed \cite{zhao2023fast}. However, SAM's reliance on manually provided prompts remains a key limitation for automated inspection tasks, such as train fault detection, where fully autonomous operation is required in practice.

Recent studies have explored adapting foundation models to specific domains while retaining their general-purpose capabilities. In industrial scenarios, achieving a balance between automation and precision is critical. For freight train inspection, key challenges include leveraging SAM's segmentation power while addressing issues related to automatic prompt generation and computational overhead. This paper tackles these challenges by introducing a prompt generator for automated prompt creation and integrating a lightweight TinyViT-SAM backbone to support efficient deployment on edge devices.

\subsection{Prompt Learning}
Prompt learning has emerged as a novel learning approach in the machine learning field in recent years, primarily through designing appropriate input prompts that enable pre-trained models to better adapt to downstream tasks without extensive fine-tuning \cite{liu2023pre}. Traditional machine learning methods, such as fully supervised learning, typically rely on large-scale labeled data and model fine-tuning to complete specific tasks, while transfer learning utilizes knowledge from source tasks to assist target tasks. However, these methods still depend on labeled data and face challenges such as high computational costs during training and insufficient data.

With the advent of large-scale pre-trained models, prompt learning has gained widespread attention. The core idea of this approach is to guide model processing through designed input prompts without directly altering the model structure or conducting extensive fine-tuning. Initially, prompt design primarily relied on manual templates, guiding models to generate expected outputs by inserting task-specific keywords or sentences. Although this method is intuitive and effective, its flexibility and scalability are limited when facing complex tasks. To overcome these limitations, researchers have proposed heuristic template generation methods and explored optimizing continuous prompting to further enhance task adaptability \cite{lester2021power}.

Building on this foundation, prompt learning has gradually developed more sophisticated strategies, such as in-context learning \cite{lei2025improving} and instruction tuning \cite{wang2024large} based on large language models. Through instruction tuning, models can execute specific tasks according to input instructions without requiring independent fine-tuning for each task, greatly improving model versatility and efficiency. Simultaneously, prompt learning has been successfully applied to computer vision, particularly in semantic segmentation and object detection tasks, where researchers have designed input prompts suitable for visual tasks, enabling pre-trained models to efficiently perform image segmentation and related operations \cite{chen2024rsprompter}.
Despite significant progress in prompt learning, several challenges remain. These include how to automatically generate effective prompts, how to ensure prompt generalization, and how to address semantic differences between various tasks \cite{sahoo2024systematic}. This paper proposes a prompter specifically designed to provide input prompts for SAM, where the input prompts are category-related and enable SAM to effectively adapt to freight train defect detection tasks.

\section{METHOD}
This section presents SAM FTI-FDet, a prompt-driven transfer learning approach designed to adapt the general segmentation knowledge of SAM to the domain of freight train image analysis. A self-prompt mechanism is introduced to enable precise domain adaptation, thereby enhancing segmentation accuracy and model robustness. As illustrated in Fig. 2, the architecture builds upon the encoder-decoder structure of SAM, with the integration of a dedicated prompt generator. Furthermore, a feature enhancement strategy is employed to mitigate the performance degradation typically associated with lightweight model designs.

\begin{figure}[!t]
    \centering
    \subfigure[Ground truth]{
    \begin{minipage}{0.28\linewidth}
			\includegraphics[width=1.1in]{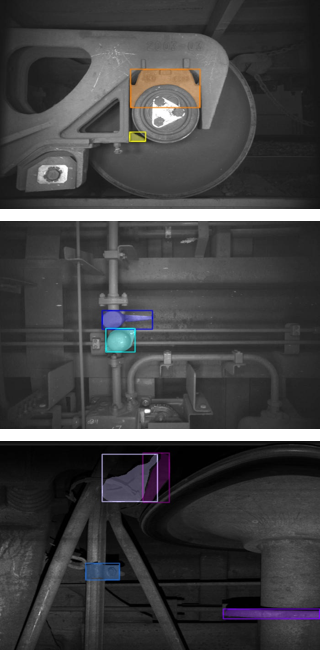}\vspace{0.4em} 
		\end{minipage}
    }
    \subfigure[Conventional Head]{
    \begin{minipage}{0.28\linewidth}
			\includegraphics[width=1.1in]{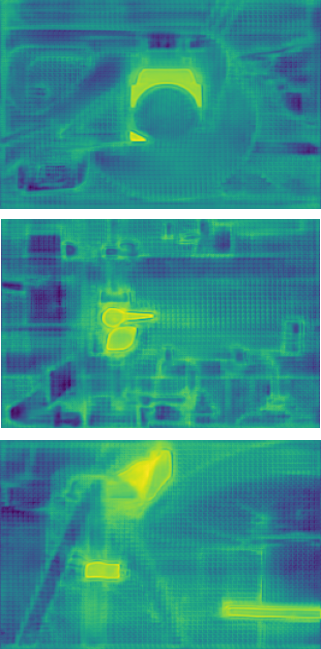}\vspace{0.4em} 
		\end{minipage}
    }
    \subfigure[Query Prompt]{
    \begin{minipage}{0.28\linewidth}
			\includegraphics[width=1.1in]{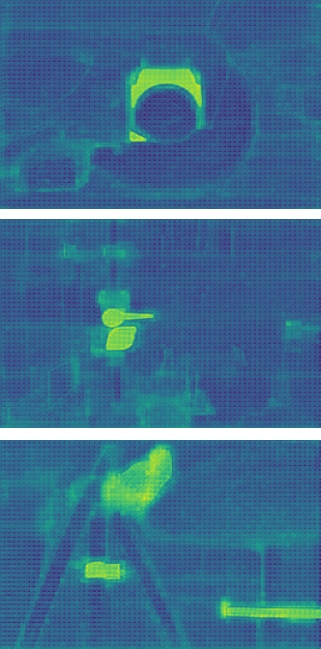}\vspace{0.4em} 
		\end{minipage}
    }
        \caption{The visualization comparison results of feature maps with and without prompts in the mask decoder stage are presented based on three typical viewpoints. (a) Ground truth annotations. (b) Feature maps generated using a conventional detection head without prompt guidance. (c) Feature maps obtained after embedding prompts in the mask decoder stage. With the proposed architecture, the target regions in (a) can be reliably detected.}
	\label{featuremap}
\end{figure}

\subsection{Overall Framework}
We propose an instance segmentation architecture based on a lightweight SAM, in which a streamlined image encoder and mask decoder are obtained through decoupled knowledge distillation from the original SAM. This approach significantly reduces computational complexity and memory overhead while preserving the segmentation quality of the original model to the greatest extent possible. It is particularly well-suited for deployment in resource-constrained environments.
The primary objective of this framework is to enhance the transferability and segmentation performance of SAM in downstream instance segmentation tasks, especially in complex industrial and transportation-related scenarios. By providing more precise prompting information and richer discriminative multi-scale feature representations, our method enables the model to focus more effectively on regions of interest. Consequently, it achieves higher segmentation accuracy and robustness under constrained computational budgets. The proposed design demonstrates outstanding performance on freight train image datasets, outperforming conventional segmentation approaches and unmodified SAM variants, particularly in scenes characterized by occlusion, background clutter, and structural complexity.

This architecture emphasizes the critical importance of accurate prompt generation and multi-scale feature fusion, both of which are central to achieving high-quality instance-level segmentation. Specifically, the data preparation process prior to the decoder stage is defined as follows:

\begin{equation}\label{query}
\begin{aligned}
 \{\bm{\mathrm{ {\hat F}}}^i\}  = {\phi _{self - attn}}({Cat{(\bm{\mathrm{F}}_{ms},\bm{\mathrm{PE}}^i,\bm{\mathrm{LE}}^i)}}),
\end{aligned}
\end{equation}
where $\bm{\mathrm{F}}_{ms}$ denotes the multi-scale image features extracted from earlier stages, while $\bm{\mathrm{PE}}^i$ and $\bm{\mathrm{LE}}^i$ represent the positional and layer encodings for the $i$-th level, respectively. The operation $Cat(\cdot)$ indicates channel-wise concatenation, and $\phi{self-attn}$ is a self-attention module designed to fuse contextual information. The resulting features ${\bm{\mathrm{\hat F}}^i}$ serve as queryable representations for subsequent decoding.

Prompts serve as crucial guiding signals within the proposed architecture, enabling the model to selectively attend to task-relevant spatial regions, thereby substantially enhancing the precision of instance segmentation. By amplifying the representation of salient objects while simultaneously suppressing background noise, the prompts effectively modulate the entire feature extraction process. In our framework, prompt signals are progressively integrated into the decoding stage via multi-scale attention mechanisms. Specifically, at each hierarchical level $i$, the query features $\bm{\mathrm{Q}}^i$ are generated based on the intermediate visual representations $\hat{\bm{\mathrm{F}}}^{i-1}$ and the queries $\bm{\mathrm{Q}}^{i-1}$ inherited from the previous level. These queries are subsequently encoded through a dedicated prompt encoding module and element-wise multiplied with the corresponding image features $\bm{\mathrm{F}}_{\text{img}}^i$, resulting in spatially modulated features. A dense embedding module then transforms the modulated features into dense prompt embeddings $\bm{\mathrm{E}}_{\text{dense}}^i$. These embeddings, together with the sparse prompts $\bm{\mathrm{E}}_{\text{sparse}}^i$, positional encodings $\bm{\mathrm{E}}_{\text{pos}}$, and image features $\bm{\mathrm{F}}_{\text{img}}^i$, are jointly fed into the cross-attention mechanism of the decoder to generate the final segmentation mask $\bm{\mathrm{M}}^i$.

This process is recursively executed across multiple scales to produce hierarchical semantic outputs. During inference, only the prediction from the final layer is retained, from which accurate object masks and bounding boxes are derived via morphological post-processing. As illustrated in Fig.~\ref{featuremap}, the resulting prompt-guided feature maps demonstrate strong localization of target regions, validating the effectiveness of the proposed strategy in enhancing both representational quality and segmentation accuracy.

\subsection{Prompt Generator}
To more effectively guide task-relevant prior information, we propose a prompt generator module based on the multi-head attention mechanism. Unlike conventional approaches such as template matching or simple linear projection, our method leverages the powerful representational capacity of multi-head attention in Transformer decoders to generate semantically rich and contextually adaptive prompt vectors through a layer-wise query refinement process. This design enhances the model’s ability to inject relevant prior knowledge and adapt to varying input distributions.

Specifically, we initialize a set of learnable query vectors $\bm{\mathrm{Q}}_0$ of length $N_q$, which serve as the foundational seeds for prompt generation. These queries are passed through a Transformer decoder composed of $L$ stacked layers, where each layer consists of two essential components: multi-head self-attention and multi-head cross-attention. This hierarchical architecture enables the progressive modeling of semantic dependencies while integrating external guidance at each stage.

In the $i$-th multi-head self-attention layer, the input query vectors $\bm{\mathrm{Q}}_{i-1}$ are first projected into distinct subspaces using the learnable matrices $\bm{\mathrm{W}}_i^Q$, $\bm{\mathrm{W}}_i^K$, and $\bm{\mathrm{W}}_i^V$ to obtain the query, key, and value representations. This subspace decomposition allows the model to capture diverse semantic patterns. The attention weights are then computed via scaled dot-product attention, followed by a softmax operation, and finally applied to a weighted sum over the value vectors. The process is formally defined as:
\begin{equation}\label{mhsa}
\begin{aligned}
\bm{\mathrm{{head}}}^s_i = softmax (\frac{{\bm{\mathrm{Q}}_{i - 1}\bm{\mathrm{W}}_i^Q{{({\bm{\mathrm{Q}}_{i - 1}}\bm{\mathrm{W}}_i^k)}^T}}}{{\sqrt {{d_k}} }})(\rm{{\bm{\mathrm{Q}}_{i - 1}}\bm{\mathrm{W}}_i^v}),
\end{aligned}
\end{equation}
where $d_k$ denotes the dimension of each attention head, serving as a scaling factor to prevent gradient explosion.

Following self-attention, the refined query vectors $\tilde{\bm{\mathrm{Q}}}_i$ are used in the multi-head cross-attention module to incorporate external features. 
Both the query and the external key/value inputs are linearly projected using matrices $\bm{\mathrm{W}}_i^{Q,c}$, $\bm{\mathrm{W}}_i^{K,c}$, and $\bm{\mathrm{W}}_i^{V,c}$, respectively. 
Dot-product attention is then applied to estimate the relevance between the internal queries and external keys. An attention mask $\bm{\mathrm{M}}$ is introduced to suppress attention to irrelevant positions, thereby enhancing selective information integration. The weighted sum over values yields the cross-attention output:
\begin{equation}\label{mhca}
\begin{aligned}
\bm{\mathrm{head}}_i^c = softmax (\frac{{{\bm{\mathrm{\tilde Q}}}_i}\bm{\mathrm{W}}_i^{Q,c}{{(\bm{\mathrm{KW}}_i^{k,c})}^T}}{{\sqrt {{d_k}} }} + \bm{\mathrm{M}}){(\bm{\mathrm{KW}}_i^{v,c})},
\end{aligned}
\end{equation}
where $\bm{\mathrm{head}}^s_i$ and $\bm{\mathrm{head}}^c_i$ denote the outputs of the $i$-th self-attention and cross-attention heads, respectively.

To obtain the updated prompt representation at each layer, the outputs from all cross-attention heads are concatenated and passed through a residual connection followed by layer normalization. This step produces the refined query vector $\bm{\mathrm{Q}}_i$ for the next layer:
\begin{equation}\label{mhca}
\begin{aligned}
Prompt &= {\bm{\mathrm{ Q}}_i} = layernorm({\bm{\mathrm{ Q}}_{i - 1}} + cat(\bm{\mathrm{head}}_i^c)) .
\end{aligned}
\end{equation}
Finally, in the subsequent mask decoding stage, the resulting prompt vector $\bm{\mathrm{Q}}_i$ serves as the query input to guide the decoding process. This enables the model to perform target-aware mask prediction, by focusing on task-specific regions informed by the prompt. The proposed design enhances the decoder’s ability to accurately localize and segment relevant objects, particularly under challenging conditions such as occlusion or background clutter.

\subsection{Adaptive Feature Dispatcher}
As shown in Fig.~\ref{structure}, the adaptive feature dispatcher is designed to achieve efficient fusion of multi-level visual features and construct task-adaptive multi-scale representations. This module fully exploits hierarchical features extracted from the SAM backbone network by employing a synergistic strategy of global aggregation and local decomposition. By doing so, it generates feature representations that are both semantically rich and detail-preserving, thereby offering robust support for subsequent vision tasks.

This module consists of two primary components: the feature aggregator and the feature splitter. The feature aggregator is responsible for the integration and dimensionality reduction of multi-layer features, whereas the feature splitter decomposes the aggregated features into multi-scale branches to adapt to varying task requirements.

Specifically, let the feature map extracted from the $i$-th layer of the backbone be denoted as ${\bm{\mathrm{F}}_i} \in \mathbb{R}^{H \times W \times C}$. Each $F_i$ is first processed by a downsampling model $\Phi_{DownConv}$.
The module $\Phi_{DownConv}$ applies a $1\times1$ convolution to reduce the channel dimension from $C$ to a fixed hidden size of 32, followed by the batch normalization and a ReLU activation. It then applies a $3\times3$ convolution with padding 1 to enrich spatial details, again followed by batch normalization and ReLU. The resulting enhanced feature is denoted as $\bm{\mathrm{\tilde F}}_i$. This design compresses the semantic content while retaining essential spatial information.

To further enhance the expressive capacity of fused features, we adopt a recursive residual aggregation strategy. Let the initial aggregated feature be defined as ${\bm{m}_i} = \bm{\mathrm{\tilde{F}}_i}$. The recursive update for subsequent layers is then given by:
\begin{equation}\label{nnorm}
\begin{aligned}
{\bm{\mathrm{m}}_i} = {\bm{\mathrm{m}}_{i - 1}} + {\Phi _{Conv2D}}({\bm{\mathrm{m}}_{i - 1}}) + {\bm{\mathrm{\tilde F}}_i},
\end{aligned}
\end{equation}
where $\Phi_{Conv2D}$ denotes a $3\times3$ convolutional layer with ReLU activation used to extract local contextual information from intermediate representations. This recursive formulation facilitates stable information flow across layers and enhances cross-scale semantic alignment.

After the recursive aggregation, $\bm{m}_i$ is passed through a fusion module $\Phi_{FusionConv}$, which refines and merges multi-layer information. This module first applies one $1\times1$ convolution followed by batch normalization and ReLU to restore the channel dimensionality, then processes the result with two consecutive $3\times3$ convolutional layers, each followed by batch normalization and ReLU, and finally applies one more $3\times3$ convolution without activation. The output of this stage is the unified aggregated feature $\bm{\mathrm{F}}_{agg}$.
Based on $\bm{\mathrm{F}}_{agg}$, the feature splitter decomposes the fused representation into multiple resolutions, enabling downstream tasks to operate effectively across different scales.

These developments highlight the increasing importance of instance segmentation in industrial fault detection. By enabling precise localization and segmentation of structural components, such as brake shoe pins, bearing saddles, and brake shoes, instance segmentation not only provides fine-grained, interpretable outputs but also allows for quantitative assessment of component conditions, such as estimating the wear level of brake shoes through area measurements and shape comparisons. This capability makes instance segmentation particularly valuable for predictive maintenance in complex and dynamic railway environments, where real-time, robust, and generalizable solutions are essential.

\subsection{Mask Decoder}
The mask decoder serves as a pivotal component in our architecture, translating the abstract semantic prompts produced by the prompt generator into concrete pixel-wise segmentation masks. Although architecturally similar to the prompt generator, as both are constructed from layered transformer blocks comprising self-attention, cross-attention, and feed-forward networks, their functional roles are fundamentally divergent. The prompt generator acts as a semantic initializer, encoding task-aware priors into a series of prompt embeddings ${\bm{\mathrm{E}}_{dense}^i \in {{\mathbb R}^{{N_P} \times C}}}$, while the mask decoder is tasked with executing the visual grounding of these prompts by integrating them with spatial image features ${\bm{\mathrm{F}}_{img}^i \in {{\mathbb R}^{H \times W \times C}}}$, ultimately producing the final segmentation mask $M \in {[0,1]^{H \times W}}$. The transformation process within the mask decoder can be formally expressed as:
\begin{equation}\label{decoder}
\begin{aligned}
{{\bm{\mathrm{H}}}^l} = \underbrace {TransBloc{k^l} \circ ... \circ TransBloc{k^1}}_{\begin{array}{*{20}{c}}
L&{Layers}
\end{array}}({{\bm{\mathrm{H}}}^0},{\bm{\mathrm{F}}}_{img}^i),
\end{aligned}
\end{equation}
where ${\bm{\mathrm{H}}^0} = \bm{\mathrm{E}}_{dense}^i$ initializes the process, and each successive transformation incorporates residual connections to facilitate gradient flow and preserve information fidelity throughout the deep architecture. As these prompt tokens propagate through the $L$ stacked transformer layers, they undergo progressive refinement, evolving into representations that simultaneously capture rich semantic context and precise spatial localization cues. The tokens comprising ${\bm{\mathrm{H}}}^l$ at the final layer individually correspond to specific prediction targets or candidate mask regions, engaging in sophisticated cross-attention operations with the multi-scale image features. This attention-driven mechanism enables the model to selectively focus on relevant visual evidence, subsequently generating highly accurate segmentation masks that delineate object boundaries with remarkable precision while maintaining semantic consistency with the original prompts.

By design, the mask decoder functions not simply as a downstream prediction head but as a prompt-sensitive executor—it listens to the semantic guidance encoded in the prompt tokens and acts upon it by precisely localizing the relevant regions in the image. This architectural arrangement grants the decoder the capacity to selectively attend to failure-related patterns, ensuring that the segmentation is both visually accurate and semantically consistent with the prompt’s intent. Its role is thus indispensable in bridging abstract task conditions and concrete spatial outputs, anchoring the high-level reasoning initiated by the prompt generator to the image domain with pixel-level precision.

\begin{figure}[!t]
	\centering	
        \includegraphics[width=3.4in]{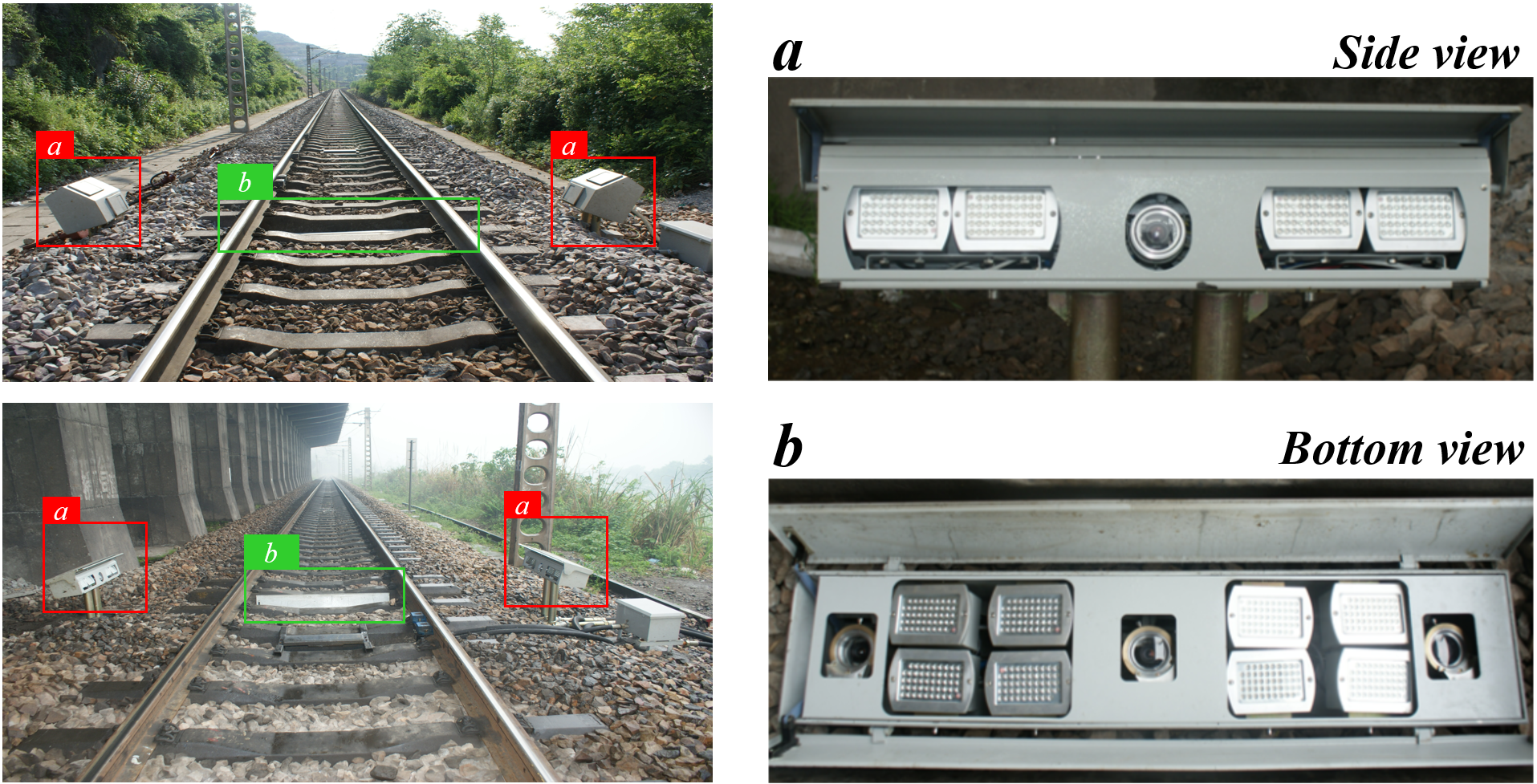}
        \caption{Image acquirement in the wild. (a) Side view. (b) Bottom view. }
	\label{hardware}
\end{figure}

\begin{figure*}
\centering
       \centering
       \includegraphics[width=6.8in]{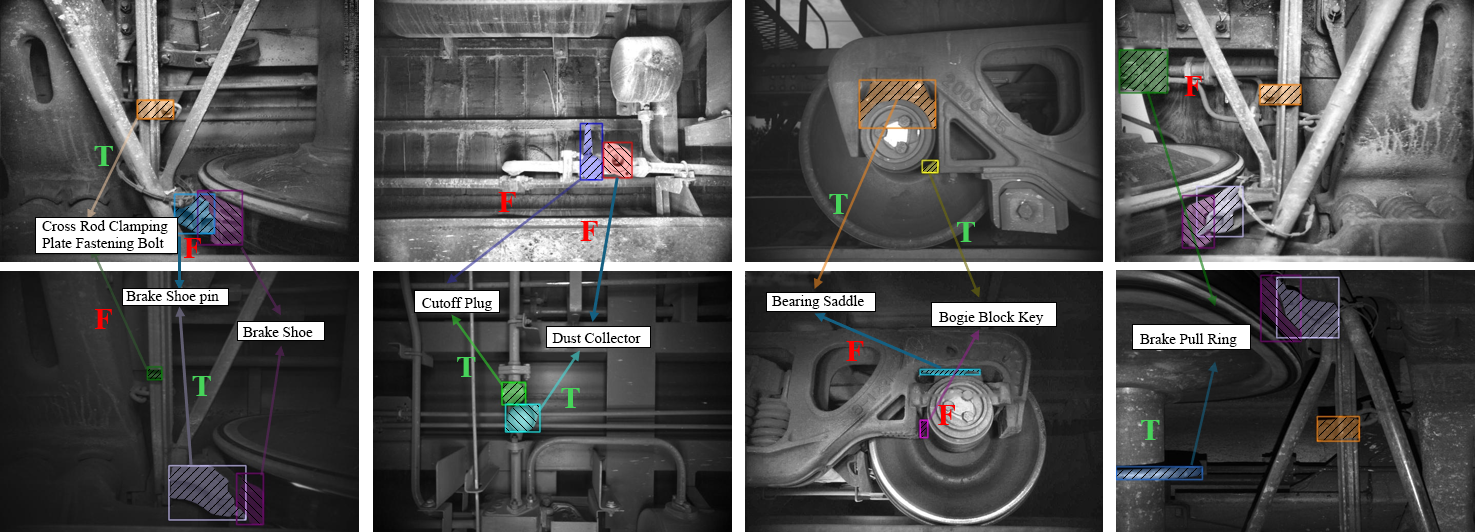}
        \caption{Visualization samples of the dataset, where “T” indicates Normal and “F” indicates Damaged or Missing. For brake shoes, the primary criterion is their thickness. For other parts, the main criterion is the presence of damage or absence.}
	\label{dataset}
\end{figure*}

\begin{table}
    \caption{Freight Train Fault Detection Dataset.}
    \label{data}
    \footnotesize
    \centering
    \renewcommand{\arraystretch}{1.2}
    \setlength{\tabcolsep}{0.55mm}
    \begin{tabular}{cccc|cccc}
        \toprule
        \multirow{2}*{Part}  &\multirow{2}*{Type}  &\multicolumn{2}{c|}{Total Images} &\multirow{2}*{Part} &\multirow{2}*{Type} &\multicolumn{2}{c}{Total Images}\\
        \cline{3-4} \cline{7-8}
                                &   &train  &val & & &train  &val\\
        \midrule
        \multirow{2}*{Bogie Block Key} &T      &725   &244 
        &\multirow{2}*{Cutoff Plug} &T      &665   &286 \\ 
                                &F &201 &101 & &F &382 &371\\ \hline
        \multirow{2}*{Brake Shoe Pin} &T      &442   &302
        &\multirow{2}*{Clamping Bolt} &T      &646   &224\\
                                &F &363 &53 & &F &157 &124\\\hline
        \multirow{2}*{Bearing Saddle} &T      &720   &244
        &\multirow{2}*{Dust Collector} &T      &443   &247\\
                                &F &201 &101 & &F &86 &98\\\hline
        \multirow{2}*{Brake Pull Ring} &T      &443   &247
        &\multirow{2}*{Brake Shoe} &\multirow{2}*{T}      &\multirow{2}*{436}   &\multirow{2}*{261}\\
                                &F &86 &98\\                  
        \bottomrule
    \end{tabular}
\end{table}

\section{EXPERIMENTS}
\subsection{Experiments Setup}
\subsubsection{Datasets}
{This study employs a freight train dataset to evaluate the model’s performance on instance segmentation tasks. The dataset consists of 4,410 images covering 6 scenarios and 15 classes. The data were collected from multiple freight train inspection stations, encompassing both side-view and bottom-view perspectives to capture diverse fault patterns. The imaging system utilizes industrial cameras (Basler acA1920 series) equipped with 16 mm C-mount lenses, capturing images under natural daylight conditions and LED-assisted illumination at night, as illustrated in Fig. \ref{hardware}. The images naturally include environmental interferences such as dust, grease, and occlusions. The images were randomly divided into training and testing sets, with 70$\%$ allocated for training and 30$\%$ for testing. Further details of the dataset, including the image resolution of 700 × 512, are provided in Table \ref{data} and Fig. \ref{dataset}.

Annotations were manually performed by railway inspectors using LabelMe to generate instance-level polygonal masks, which were subsequently reviewed by a senior expert to ensure consistency and accuracy. For wear-related components such as brake shoes, the severity levels were defined according to the remaining usable material thickness: slight wear (70\%–100\%), moderate wear (30\%–70\%), and severe wear ($<$30\%). For example, given a brake shoe with an original thickness of 45 mm and a usable wear allowance of 20 mm, a remaining thickness below 25 mm was classified as unserviceable. For structural components (e.g., Bogie Block Key, Brake Shoe Pin), faults were categorized as normal or damaged. The fault definitions were established in accordance with the Railway Freight Car Operation and Maintenance Code (TB 10057–2021) in China.}

{\color{blue}In addition, the MS COCO \cite{lin2014microsoft} dataset was also used to further validate the effectiveness of the proposed method through independent fine-tuning.}

\begin{table*}[t]
    \renewcommand{\arraystretch}{1.25}
    \caption{Comparison of accuracy with the state-of-the-art methods on the freight trains dataset.}
    \label{SOTA}
    \setlength{\tabcolsep}{1.1mm}
    \centering
    \begin{tabular}{c|lc|ccc|ccc|cccc}
        \toprule
        \multicolumn{2}{c}{{\textbf{Method}}} &{\textbf{Backbone}} &{$AP^{box}$} &{$AP^{box}_{50}$} &{$AR^{box}_{50-95}$}& {$AP^{mask}$} &{$AP^{mask}_{50}$} &{\textbf{$AR^{mask}_{50-95}$}} &{Model Size} &GFLOPs &FPS &Params\\
        \midrule
        \multirow{9}[1]{*}{\begin{turn}{90}\emph{\textbf{CNN}}\end{turn}}
        &Mask R-CNN\cite{he2017mask} &ResNet50 &70.1 &94.8 &78.2 &70.7 &93.7 &77.9 &336.4 &234 &44.6 &44.0 \\
        &PointRend\cite{kirillov2020pointrend} &ResNet50 &71.3 &94.2 &78.4 &71.3 &94.2 &78.4 &431.6 &186 &38.1 &56.4\\
        &YOLACT\cite{bolya2019yolact} &ResNet50 &63.6 &93.0 &71.2 &68.4 &93.0 &76.0 &272.0 &337 &42.8 &31.4\\
        &SOLOv2\cite{wang2020solov2} &ResNet50 &-- &-- &-- &69.8 &93.8 &76.8 &353.9 &283 &33.7 &46.3\\
        &CondInst\cite{tian2020conditional} &ResNet50 &69.4 &93.2 &74.7 &67.3 &92.8 &73.3 &272.0 &250 &27.1 &48.2\\
        &SparseInst\cite{Cheng2022SparseInst} &ResNet50 &-- &-- &-- &66.5 &87.9 &72.2 &399.8 &221 &\textbf{76.3} &47.9\\
        &RTMDet\cite{lyu2212rtmdet} &CSPNeXt\cite{chen2024cspnext} &71.3 &94.2 &61.2 &68.8 &94.0 &74.8 &436.0 &\textbf{120} &23.1 &32.5\\
        &QueryInst\cite{Fang_2021_ICCV} &ResNet50 &73.0 &77.4 &\textbf{82.6} &66.8 &94.5 &77.5 &1986.6 &263 &16.0 &62.5\\
        &Mask2Former\cite{cheng2022masked} &ResNet50 &74.2 &92.7 &80.1 &72.6 &93.3 &78.9 &665.2 &245 &13.0 &46.3\\
        \midrule
        \multirow{7}[1]{*}{\begin{turn}{90}\emph{\textbf{Transformer}}\end{turn}}
        &Mask R-CNN\cite{he2017mask} &Swin-T\cite{liu2021Swin} &72.6 &\textbf{95.4} &78.9 &70.5 &95.3 &76.8 &549.0 &241 &36.4 &47.5\\
        &PointRend\cite{kirillov2020pointrend} &Swin-T\cite{liu2021Swin} &69.0 &95.1 &76.6 &71.8 &95.0 &\textbf{79.1} &688.2 &198 &30.1 &65.3\\
        &YOLACT\cite{bolya2019yolact} &Swin-T\cite{liu2021Swin} &59.9 &82.2 &67.8 &66.6 &92.7 &74.3 &275.4 &282 &31.6 &50.2\\
        &SOLOv2\cite{wang2020solov2} &Swin-T\cite{liu2021Swin} &-- &-- &-- &67.0 &94.4 &75.8 &380.6 &312 &28.3 &49.7\\
        &CondInst\cite{tian2020conditional} &Swin-T\cite{liu2021Swin} &65.6 &94.5 &75.2 &72.6 &94.9 &78.8 &293.4 &258 &30.6 &68.3 \\
        &RTMDet\cite{lyu2212rtmdet}  &Swin-T\cite{liu2021Swin} &73.5 &94.4 &79.3 &70.3 &94.4 &75.3 &1790.6 &323 &25.6 & 86.6\\
        &Mask2Former\cite{cheng2022masked} &Swin-T\cite{liu2021Swin} &74.3 &93.0 &81.0 &73.8 &93.0 &79.3 &739.5 &252 &12.8 &49\\
        \midrule
        \multirow{6}[1]{*}{\begin{turn}{90}\emph{\textbf{SAM}}\end{turn}}
        &FastSAM\cite{zhao2023fast} &CSPDarknet\cite{wang2020cspnet} &73.6 &92.0 &82.1 &72.0 &\textbf{95.1} &77.9 &141.6 &443 &9.1 &72 \\
        &SAM-seg \cite{chen2024rsprompter} &SAM-B\cite{kirillov2023segment} &72.4 &94.6 &78.9 &71.9 &94.7 &77.4 &450.2 &411 &8.6 &97.2 \\
        &SAM-det \cite{chen2024rsprompter} &SAM-B\cite{kirillov2023segment} &72.0 &93.2 &79.0 &58.9 &84.5 &67.9 &528.0 &233 &10.2 &106.7\\
        &RSPrompter-query \cite{chen2024rsprompter} &SAM-B\cite{kirillov2023segment} &72.7 &92.7 &79.7 &71.9 &93.2 &77.8 &303.3 &425 &7.1 &131\\
        &\textbf{SAM FTI-FDet-PF} &TinyVit-SAM &73.2 &93.4 &81.2 &72.9 &93.8 &79.5 &\textbf{80.4} &196 &24.4 &\textbf{30.1} \\
        &\textbf{SAM FTI-FDet} &TinyVit-SAM &\textbf{74.6} &93.9 &81.4 &\textbf{74.2} &94.8 &\textbf{79.9} &148.2 &244 &16.0 &36.3\\
        \bottomrule
    \end{tabular}
\end{table*}

\subsubsection{Evaluation Metrics}
{In this experiment, we use a set of evaluation metrics to assess both the accuracy and complexity of the model. The metrics $AP^{box}$ and $AP^{mask}$ represent the mean average precision for bounding box detection and instance segmentation respectively, calculated across IoU thresholds from 0.50 to 0.95. $AP^{box}_{50}$ and $AP^{mask}_{50}$ refer to the average precision at an IoU threshold of 0.50, reflecting the model’s performance under a more relaxed condition. $AR^{box}_{50-90}$ and $AR^{mask}_{50-95}$ measure the average recall across multiple IoU thresholds, indicating the model’s ability to detect objects comprehensively. In addition, we report the inference speed in terms of Frames Per Second (FPS), the computational cost measured by giga floating-point operations (GFLOPs), together with the number of model parameters (Params) and the overall model size (MB) to evaluate the computational complexity and deployment feasibility.}

\subsubsection{Implementation Details}
{All experiments were conducted on the aforementioned datasets using consistent training protocols. During training, input images were uniformly resized to 1024 × 1024 pixels, in accordance with the original input specification of the SAM model. To improve dataset diversity and model generalization, we applied standard data augmentation techniques, including horizontal flipping and large-scale jittering. During inference, the model was configured to predict up to 10 instance masks per image.

For each image, we generated 10 sets of prompts ($N_p$ = 10), with each set containing 4 prompt embeddings ($K_p$ = 4). In the feature aggregation module, features were extracted from all layers of the backbone network to fully exploit representational capacity, which is especially important given the lightweight design of our backbone. In the prompt generator, we selectively used only the final three feature maps with the smallest spatial resolutions from the feature segmentation module, thereby balancing computational efficiency and information preservation in practice.

All experiments were conducted on a system equipped with dual NVIDIA GeForce RTX 4090 (24GB) GPUs. Model optimization was carried out using the AdamW optimizer with an initial learning rate of $1\times10^{-4}$ and a batch size of 4. The training process spanned 150 epochs, with the learning rate scheduled using cosine annealing and a linear warmup strategy throughout training.

Our method was implemented using the PyTorch framework, with all auxiliary modules trained from scratch. To further enhance computational efficiency in the SAM FTI-FDet setting, we employed DeepSpeed \cite{rasley2020deepspeed} ZeRO Stage 2 optimization in conjunction with FP16 mixed-precision training techniques for optimal performance.}

\subsection{Comparison With State-of-the-Art Methods}
\begin{figure*}[t]
        \subfigure[Ground Truth]{
    \begin{minipage}[c]{0.12\linewidth}
			\includegraphics[width=1in]{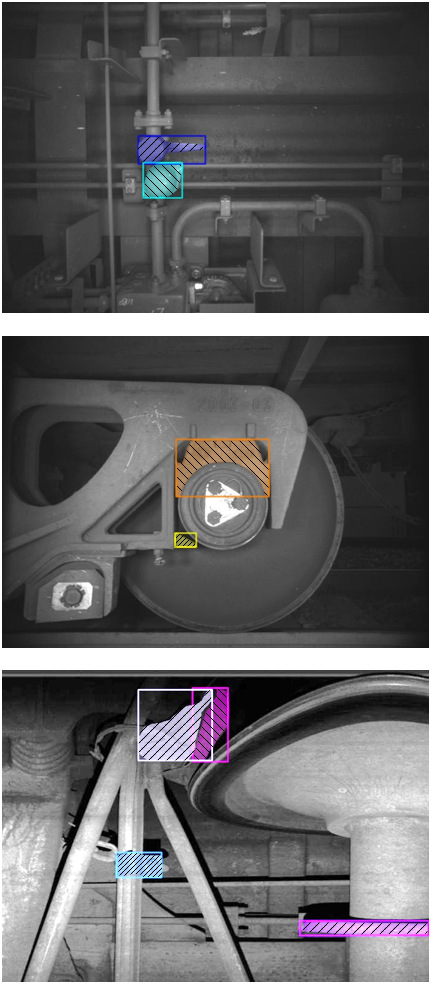}\vspace{0.4em} 
		\end{minipage}
    }\hspace{0.1em}
    \subfigure[Mask R-CNN]{
    \begin{minipage}[c]{0.12\linewidth}
			\includegraphics[width=1in]{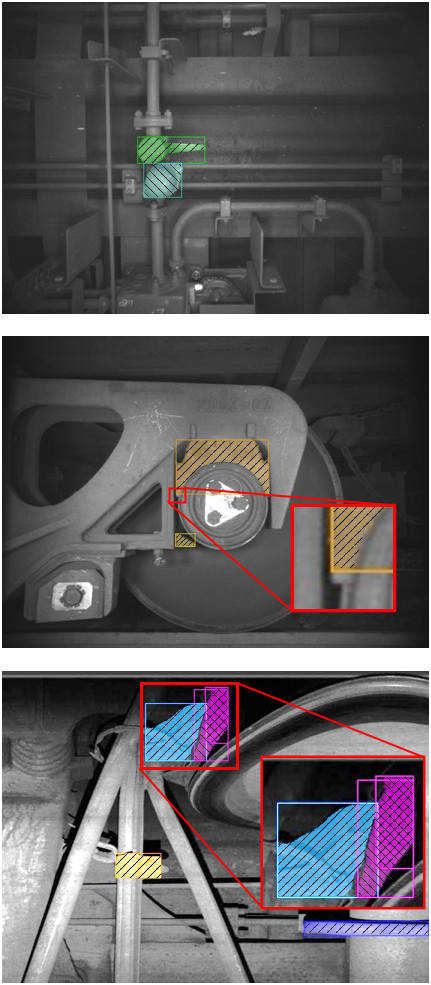}\vspace{0.4em} 
		\end{minipage}
    }\hspace{0.1em}
    \subfigure[SOLOv2]{
    \begin{minipage}[c]{0.12\linewidth}
			\includegraphics[width=1in]{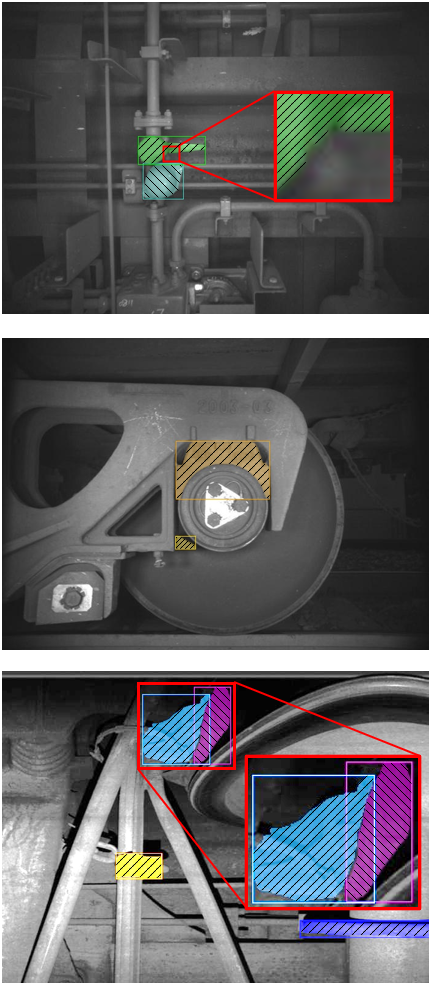}\vspace{0.4em} 
		\end{minipage}
    }\hspace{0.1em}
    \subfigure[Point Rend]{
    \begin{minipage}[c]{0.12\linewidth}
			\includegraphics[width=1in]{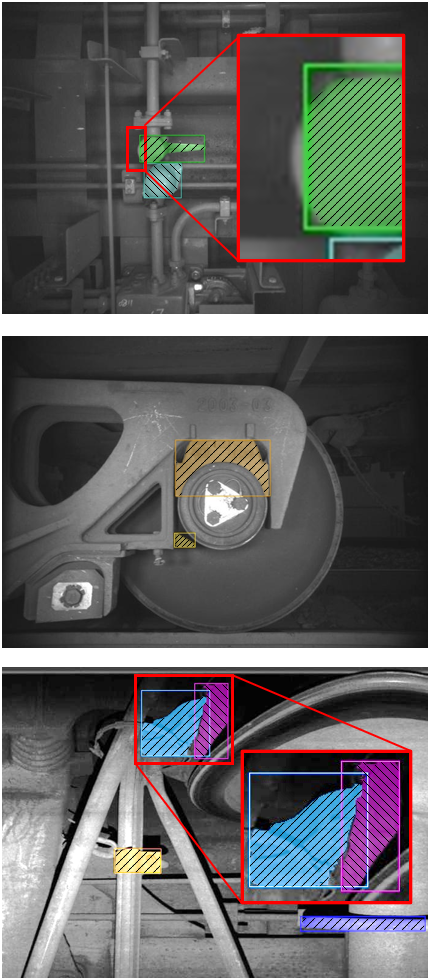}\vspace{0.4em} 
		\end{minipage}
    }\hspace{0.1em}
    \subfigure[Point Rend*]{
    \begin{minipage}[c]{0.12\linewidth}
			\includegraphics[width=1in]{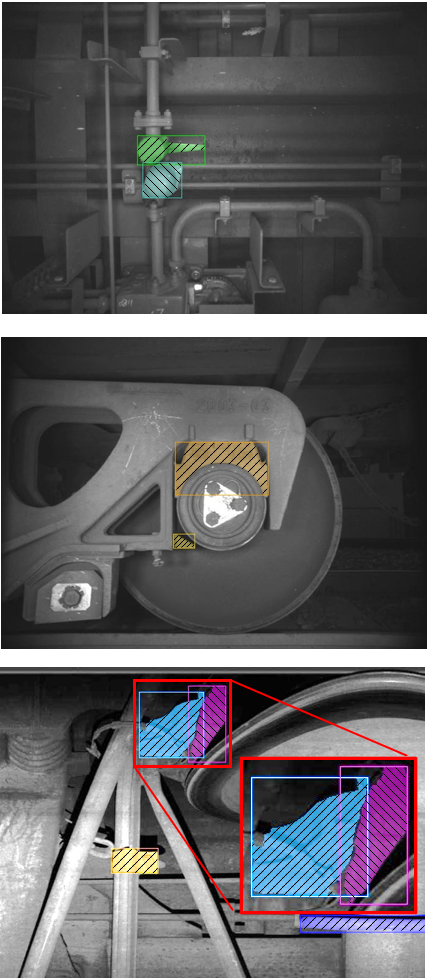}\vspace{0.4em} 
		\end{minipage}
    }\hspace{0.1em}
    \subfigure[Mask2Former*]{
    \begin{minipage}[c]{0.12\linewidth}
			\includegraphics[width=1in]{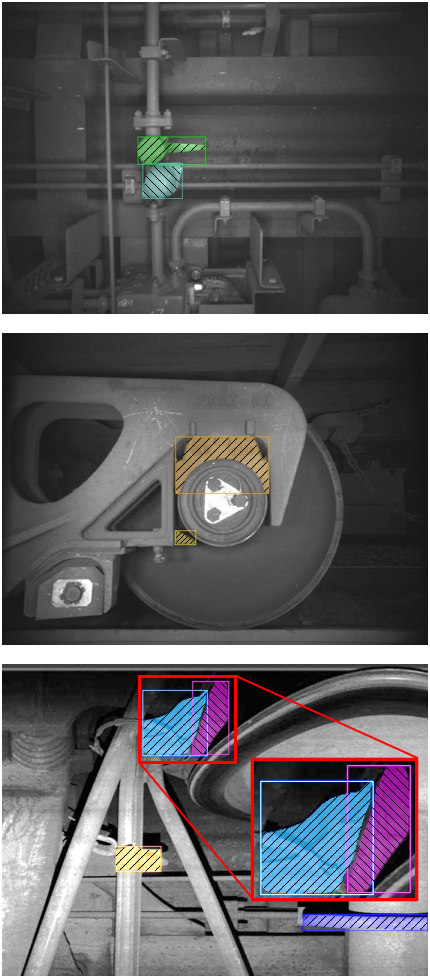}\vspace{0.4em} 
		\end{minipage}
    }\hspace{0.1em}
    \subfigure[SAM FTI-FDet]{
    \begin{minipage}[c]{0.12\linewidth}
			\includegraphics[width=1in]{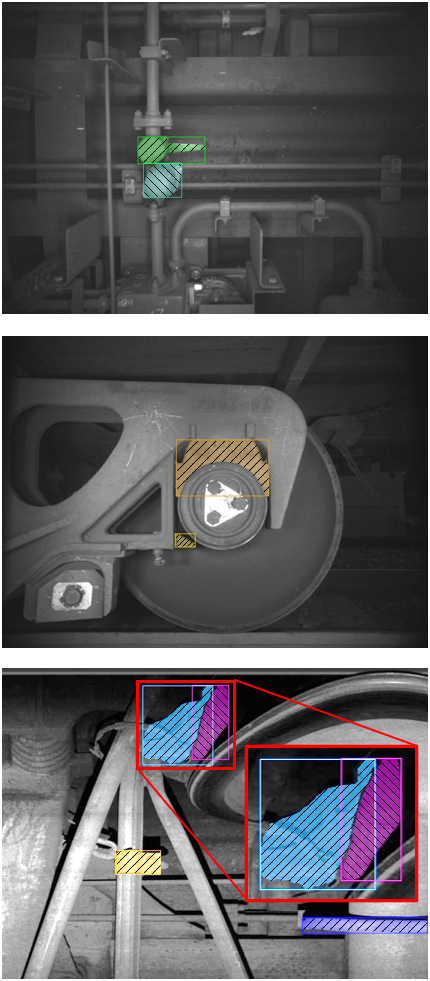}\vspace{0.4em} 
		\end{minipage}
    }
        \caption{Comparative visualization of segmentation results for image samples. The “ * ” denotes the Swin-T architecture. Evidently, the results produced by SAM FTI-FDet exhibit the closest alignment with the ground truth annotations. This advantage primarily stems from the proposed efficient prompting and knowledge transfer strategy, which substantially enhances the model’s capability to capture instance-specific semantics.}
	\label{result}
\end{figure*}

To evaluate the performance of our method, we conducted a comprehensive comparative analysis on a freight train dataset against state-of-the-art segmentation approaches. Specifically, we compared our framework with two distinct categories: traditional two-stage frameworks based on CNNs and Transformers, as well as existing SAM-based methodologies. Fig. \ref{result} presents qualitative comparison results, illustrating segmentation performance on several representative examples. As detailed in Table \ref{SOTA}, our analysis systematically assesses detection accuracy, segmentation precision, and model complexity, thereby demonstrating both the effectiveness and computational efficiency of our proposed method.

\subsubsection{Classic Segmentation}
{The experimental results highlight the strengths of different methods across specific evaluation metrics, reflecting the impact of their architectural designs. QueryInst achieves the highest $AR^{box}_{50-90}=82.6$, attributed to its query-based object modeling, which enhances recall in multi-object scenarios. Mask R-CNN with a Swin-T backbone attains the highest $AP^{box}_{50}= 95.4$, demonstrating the advantage of global receptive fields in accurately localizing objects under complex backgrounds. PointRend achieves the highest mask AP50, owing to its point-wise refinement strategy that enables superior boundary modeling.}

\subsubsection{SAM-based Segmentation}
{The SAM benefits from large-scale pretraining, enabling strong boundary modeling and semantic understanding. This allows it to outperform CNN and Transformer-based methods in complex segmentation tasks. Our prompt-free SAM FTI-FDet-PF leverages lightweight design and efficient encoding to achieve competitive performance with the lowest parameter count and minimal model size among all compared methods. With prompt information incorporated, the full SAM FTI-FDet further improves segmentation and detection performance. This improvement is attributed to the proposed query-based interaction mechanism, which enhances semantic-spatial feature alignment and improves instance awareness. This is particularly beneficial in freight train inspection, where diverse object categories, structural complexity, and small defects require precise, instance-level perception. Notably, SAM FTI-FDet achieves superior accuracy in instance-level fault detection while maintaining the lowest parameter count and computational cost among foundation models, making it highly suitable for deployment on edge-side track inspection devices with real-time, high-precision requirements. Unlike RSPrompter, which struggles in structurally complex and defect-focused train scenarios, our method balances accuracy and efficiency, making it ideal for industrial visual inspection. As shown in Fig. \ref{result}, it delivers precise boundaries and contour fidelity close to ground truth.}

\subsection{Ablation Study}
\subsubsection{Prompt evaluation}
In complex freight train detection tasks, traditional box-based prompt mechanisms often fail to capture precise object boundaries due to their coarse spatial constraints, which limits the upper bound of segmentation accuracy. In contrast, our method introduces a query-based feature interaction mechanism that leverages learnable query vectors to encode target-specific semantic priors. This design enhances the alignment between image features and task semantics, enabling more discriminative object modeling even under challenging industrial conditions such as low-quality backgrounds, occlusions, and inter-class interference.

\begin{figure}[!t]
	\centering	
        \includegraphics[width=3.4in]{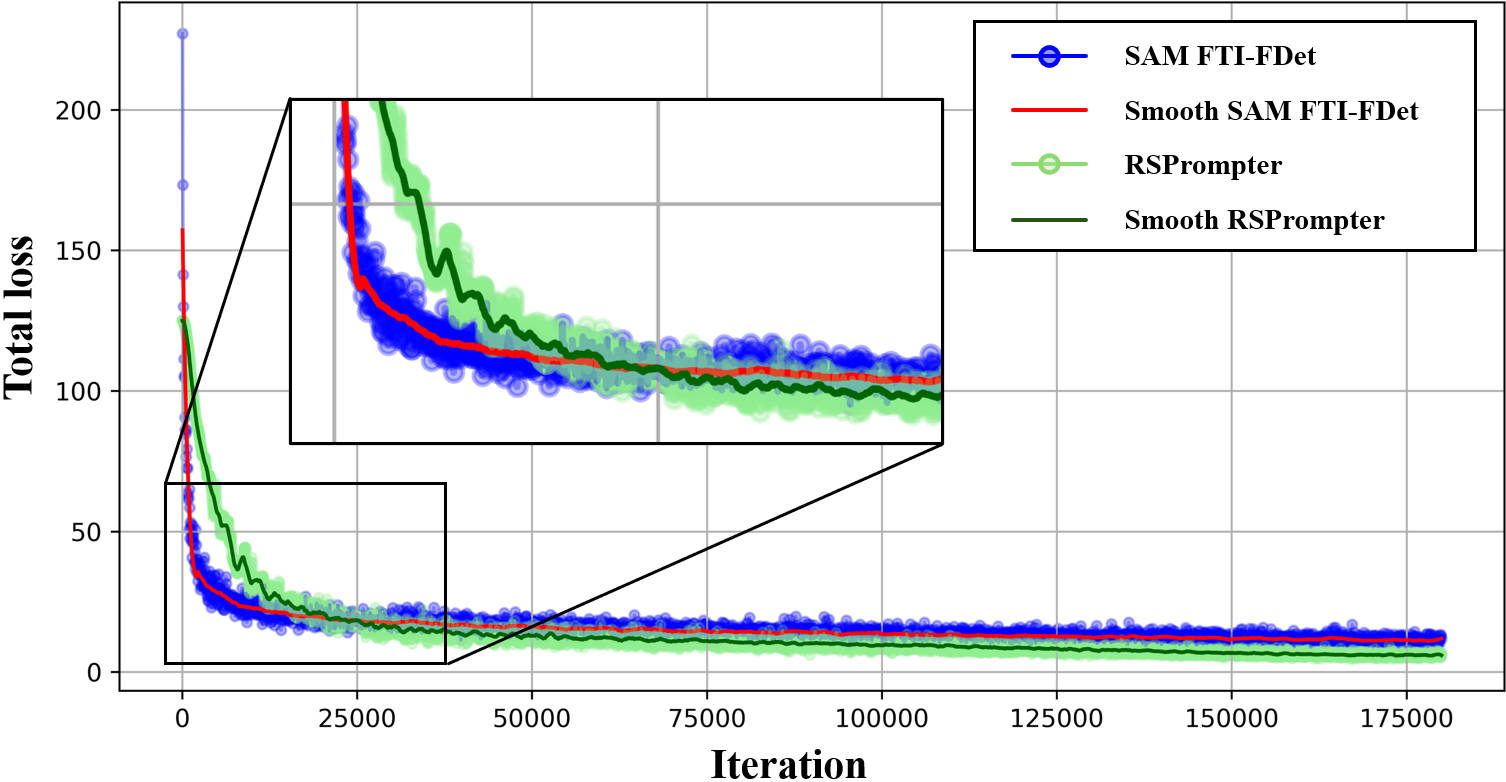}
        \caption{Comparison of total training loss between SAM FTI-FDet and RSPrompter. The proposed method exhibits markedly faster convergence, indicating that the self-prompt mechanism provides more efficient and semantically aligned guidance during optimization. This superior convergence behavior highlights the higher prompt-generation efficiency of our approach compared with query-based prompting.}
	\label{loss}
\end{figure}

Beyond semantic advantages, our prompting strategy also exhibits significantly higher optimization efficiency. As shown in Fig. \ref{loss}, the total training loss of SAM FTI-FDet decreases much faster than that of RSPrompter. This accelerated convergence empirically demonstrates that our self-prompt mechanism provides clearer, more informative guidance during optimization. Compared with RSPrompter’s query-based prompting, which involves multiple manually designed transformations, our two-way Transformer prompt generator produces prompts in a more direct, compact, and semantically aligned manner. The improved convergence behavior indicates that the model reaches a stable optimum more efficiently, quantitatively validating the effectiveness and computational superiority of our prompting design.

Furthermore, the flexibility and scalability of query prompts allow the model to dynamically adapt to different component types and varying scene conditions. This effectively overcomes the dependency of traditional prompts on predefined target regions and improves robustness and generalization in real-world inspection scenarios. As summarized in Table \ref{Prompt evaluation}, our method achieves superior detection and segmentation performance across all metrics on the freight train dataset, confirming both the semantic effectiveness and optimization efficiency of the proposed prompting strategy.

\begin{table}[t]
	\renewcommand{\arraystretch}{1.2}
	\caption{Prompt evaluation on the freight trains dataset.}
        \label{Prompt evaluation}
	\centering
	\footnotesize
	\setlength{\tabcolsep}{0.52mm}{
		\begin{tabular}{cccccc}
			\toprule
            \textbf{Method}& 
            \textbf{Prompt Type}& 
            {$AP^{box}$}&
            {$AP^{box}_{50}$}& 
            {$AP^{mask}$}&
            {$AP^{mask}_{50}$}\\
			\midrule
            SAM \cite{kirillov2023segment} & gd-bbx &-- &-- &66.3 &91.5 \\
            MobileSAM \cite{zhang2023faster} & gd-bbx &-- &-- &64.2 &90.9 \\
            FastSAM \cite{zhao2023fast} & gd-bbx &-- &-- &63.0 &89.8 \\
            SAM-det \cite{chen2024rsprompter}& bbox &72.0 &93.2 &57.7 &73.2\\
            RSPrompter \cite{chen2024rsprompter}& anchor &67.1 &93.2 &68.4 &92.6\\
            RSPrompter \cite{chen2024rsprompter} & query &72.7 &92.7 &71.9 &93.2\\
            \textbf{SAM FTI-FDet}& query &\textbf{74.6} &\textbf{93.9} &\textbf{74.2} &\textbf{94.8}\\
			\bottomrule
	\end{tabular}}
\end{table}

\begin{table}[t]
	\renewcommand{\arraystretch}{1.2}
	\caption{Decoder-head evaluation on the freight trains dataset.}
        \label{Prompt-free evaluation}
	\centering
	\footnotesize
	\setlength{\tabcolsep}{0.7mm}{
		\begin{tabular}{cccccc}
			\toprule
                \textbf{Method}&
            {\textbf{Head}}& 
            {\textbf{$AP^{box}$}}&
            {\textbf{$AP^{box}_{50}$}}& 
            {\textbf{$AP^{mask}$}}&
            {\textbf{$AP^{mask}_{50}$}}\\
			\midrule
            SAM & Mask R-CNN &72.5 &94.7 &71.8 &\textbf{94.7} \\
            SAM & Mask2Former &72.9 &92.9 &72.1 &94.0 \\
            \textbf{SAM FTI-FDet-PF} & Mask R-CNN &70.9 &\textbf{94.9} &72.1 &\textbf{94.7}\\
            \textbf{SAM FTI-FDet-PF} & Mask2Former &\textbf{73.2} &93.4 &\textbf{72.9} &93.8\\
			\bottomrule
	\end{tabular}}
\end{table}

\begin{table}[t]
	\renewcommand{\arraystretch}{1.2}
	\caption{Comparison of backbones and pretrained models on the freight train dataset.}
        \label{backbone}
	\centering
	\footnotesize
	\setlength{\tabcolsep}{0.7mm}{
		\begin{tabular}{lc|cccccc}
			\toprule
                {\textbf{Backbone}} &\textbf{Pretrain}& 
            {\textbf{$AP^{box}$}}& 
            {\textbf{$AP^{box}_{50}$}}&
            {\textbf{$AP^{mask}$}}&
            {\textbf{$AP^{mask}_{50}$}}& 
            Params\\
			\midrule
            ResNet18& ImageNet& 68.5 &93.3 &68.8 &93.4 &11M \\
            ResNet50& ImageNet&70.0 &93.7 &70.8 &94.0 &26M\\
            ResNet101&ImageNet& 70.7 &\textbf{94.0} &70.3 &93.8 &45M\\
            Swin-T& MS-COCO&73.2 &93.1 &73.2 &94.2 &28M\\
            SAM-B & SA-1B &73.6 &93.2 &73.4 &93.7 &86M\\
            TinyViT-5m & SA-1B &\textbf{74.6} &93.9 &\textbf{74.2} &94.8 &5M\\
            TinyViT-11m & SA-1B &73.5 &93.8 &73.2 &93.8 &11M\\
            TinyViT-21m & SA-1B &74.3 &\textbf{94.0} &\textbf{74.2} &\textbf{94.9} &21M \\
			\bottomrule
	\end{tabular}}
\end{table}

\begin{table}[t]
    \renewcommand{\arraystretch}{1.2}
    \caption{Segmentation performance with different levels of features fed into the aggregator, where [a, b, c] denotes the indices of the retained feature map levels.}
    \label{selectlayer}
    \centering
    \footnotesize
    \setlength{\tabcolsep}{0.5mm}{
		\begin{tabular}{c|ccc|ccc}
            \toprule
            \textbf{Select Layers}& $AP^{box}$ &$AP^{box}_{50}$ & $AR^{box}_{50-90}$ &$AP^{mask}$ &$AP^{mask}_{50}$ & $AR^{mask}_{50-90}$ \\
            \midrule
            $[0,1,2,3]$&73.8 &92.2 &80.1 &73.2 &93.3 &79.2\\
            $[1,2,3]$&74.1 &92.7 &80.9 &73.6 &93.7 &79.4\\
            $[1,3]$ &73.6 &92.7 &80.5 &73.8 &93.9 &79.5 \\
            \textbf{$[\textbf{2},\textbf{3}]$}&\textbf{74.6} &\textbf{93.9} &\textbf{81.4} &\textbf{74.2} &\textbf{94.8} &\textbf{79.9}\\
            $[3]$ &72.6 &93.0 &79.6 &72.6 &93.8 &79.6\\
            \bottomrule
            \end{tabular}}
\end{table}

\subsubsection{Decoder-head evaluation}
To more accurately evaluate the transferability of different SAM encoders, we conduct a decoder-head evaluation, where the original SAM mask decoder is replaced with standard detection and segmentation heads such as Mask R-CNN and Mask2Former. This setting enables us to examine how well SAM-style visual encoders adapt to downstream tasks without relying on SAM’s prompt-guided decoding module.
As shown in Table \ref{Prompt-free evaluation}, we compare two encoder variants: SAM-seg with the SAM-B encoder and our lightweight SAM FTI-FDet-PF with the TinyViT-SAM encoder, under both detection and segmentation frameworks. Despite its significantly reduced model size, SAM FTI-FDet-PF achieves comparable or even superior performance relative to SAM-seg, indicating that the TinyViT-based encoder retains the essential visual representation capabilities of the original SAM and transfers effectively to downstream detection tasks.

Furthermore, Mask2Former consistently outperforms Mask R-CNN, owing to its Transformer-based decoder that better models long-range dependencies and effectively handles the complex structures, occlusions, and fine-grained boundaries commonly found in freight train components. This demonstrates that the semantic priors encoded by SAM are more complementary to Transformer-style decoder heads.

\subsubsection{Impact of backbone and pretrain}
We explored various backbone networks and pre-trained models to assess their impact on performance. As shown in Table \ref{backbone}, models pre-trained on the SA-1B dataset demonstrated superior results. Although SAM-B achieved good performance, its parameter count significantly exceeds that of the TinyViT-SAM series. While TinyViT-5m slightly increases GFLOPs compared to SAM FTI-FDet-PF, it achieves notably higher $AP^{box}$ and $AP^{mask}$, substantially improving fault detection accuracy. This increase in computational cost remains within the real-time processing capacity of edge devices such as the Jetson Orin NX 16GB, making it an acceptable trade-off for practical deployment scenarios.

\subsubsection{Impact of different layer in feature aggregator}
The inputs to the feature enhancer are derived from different hierarchical features of the TinyViT backbone network. Since TinyViT consists of only four layers, we conducted a comprehensive experimental study on various layer combinations to investigate their impact on segmentation performance. As shown in Table \ref{selectlayer}, the experimental results demonstrate that using features from the final two layers [2, 3] yields optimal detection accuracy in practice.

\begin{table}[t]
    \renewcommand{\arraystretch}{1.2}
    \caption{Comparisons of frozen different part.}
    \label{frozen}
    \centering
    \footnotesize
    \setlength{\tabcolsep}{1mm}{
		\begin{tabular}{cc|cccccc}
            \toprule
            \textbf{En}& \textbf{De}& $AP^{box}$ &$AP^{box}_{50}$ & $AR^{box}_{50-90}$ &$AP^{mask}$ &$AP^{mask}_{50}$ & $AR^{mask}_{50-90}$ \\
            \midrule
            \textit{f} & \textit{f} &66.9 &90.9 &71.4 &65.7 &91.1 &71.0\\
            \textit{uf} &\textit{f}  &\textbf{74.6} &\textbf{93.9} &\textbf{81.4} &\textbf{74.2} &\textbf{94.8} &\textbf{79.9} \\
            \textit{f} &\textit{uf}  &71.4 &92.8 &79.2 &70.8 &93.0 &78.6\\
            \textit{uf} &\textit{uf} &72.2 &91.9 &79.7 &72.2 &91.9 &79.0 \\
            \bottomrule
            \end{tabular}}
\end{table}

\subsubsection{Impact of freezing mask encoder and mask decoder}
As shown in Table \ref{frozen}, the superior performance of the uf/f ('f' = frozen, 'uf' = unfrozen) configuration can be attributed to a balance between adaptability and generalization. By fine-tuning the encoder, the model learns task-specific representations that are tailored to the target domain. Meanwhile, freezing the decoder acts as a form of regularization, helping preserve the general semantic decoding structure and reducing the risk of overfitting.

\begin{table}[t]
	\renewcommand{\arraystretch}{1.2}
        \caption{Comparisons of different channels in feature aggregator and prompt generator.}
        \label{channal}
	\centering
	\footnotesize
	\setlength{\tabcolsep}{0.8mm}{
		\begin{tabular}{c|ccc|ccc}
            \toprule
            \textbf{Channels}& $AP^{box}$ &$AP^{box}_{50}$ & $AR^{box}_{50-90}$ &$AP^{mask}$ &$AP^{mask}_{50}$ & $AR^{mask}_{50-90}$ \\
            \midrule
            64 &67.3 &90.4 &76.2 &67.5 &92.1 &76.3\\
            128 &70.9 &91.7 &78.7 &71.5 &93.8 &77.5 \\
            \textbf{256} &\textbf{74.6} &\textbf{93.9} &\textbf{81.4} &\textbf{74.2} &\textbf{94.8} &\textbf{79.9} \\
            \bottomrule
            \end{tabular}}
\end{table}

\subsubsection{Impact of Feature Channels}
To investigate the impact of channel width on model performance, we conducted a set of controlled experiments by varying the number of channels to 64, 128, and 256, while keeping all other architectural parameters unchanged. The results demonstrate a clear performance improvement as the channel number increases, which is shown in Table \ref{channal}. Specifically, the configuration with 256 channels consistently outperformed the 64 and 128-channel variants across all evaluation metrics. This indicates that a higher channel width enables the model to extract richer and more discriminative features, leading to improved fault detection accuracy. However, increasing the number of channels also introduces additional computational and memory overhead. Considering the trade-off between performance and resource consumption, we ultimately selected 256 channels as the final configuration for our model.

\begin{table}[t]
    \renewcommand{\arraystretch}{1.2}
    \caption{Segmentation performance with varying query numbers and prompt embedding numbers.}
    \centering
    \label{promptshape}
    \footnotesize
    \setlength{\tabcolsep}{1mm}{
		\begin{tabular}{cc|cccccc}
            \toprule
            \textbf{$N_q$} & \textbf{$N_p$}& $AP^{box}$ &$AP^{box}_{50}$ & $AR^{box}_{50-90}$ &$AP^{mask}$ &$AP^{mask}_{50}$ & $AR^{mask}_{50-90}$ \\
            \midrule
            30 & 4 &73.2 &92.1 &\textbf{82.1} &72.8 &93.7 &79.9\\
            20 & 4 &73.7 &93.3 &80.9 &73.2 &94.6 &78.6\\
            \textbf{10} & \textbf{4} &\textbf{74.6} &\textbf{93.9} &81.4 &\textbf{74.2} &\textbf{94.8} &\textbf{79.9}\\
            5 & 4 &72.2 &91.8 &79.6 &72.7 &93.3 &78.6\\
            1 & 4 &63.0 &79.0 &66.9 &62.4 &79.9 &67.1\\
            \midrule
            10 & 10 &72.9 &92.8 &80.7 &72.8 &93.4 &79.0 \\
            10 & 8 &73.5 &93.0 &80.6 &73.2 &94.0 &79.1\\
            10 & 6 &73.7 &92.9 &80.6 &73.3 &94.1 &78.9 \\
            10 & 2 &73.2 &93.3 &80.3 &73.6 &93.8 &79.3\\
            10 & 1 &73.7 &93.4 &77.6 &73.2 &93.7 &78.7\\

            \bottomrule
            \end{tabular}}
\end{table}

\subsubsection{Impact of Prompt Shape}
We conducted a comprehensive study on the shape-related parameters of the prompt generator, namely the number of prompt groups ($N_q$) and the number of point prompts per group ($N_p$), to understand their effects on segmentation accuracy and computational costs. As shown in Table~\ref{promptshape}, performance increases steadily as $N_q$ approaches the typical number of instances in an image. This is because a larger number of prompt groups provides better instance-level coverage. However, when $N_q$ becomes too small (e.g., $N_q = 1$), performance degrades significantly due to insufficient coverage, while setting $N_q$ exactly equal to the instance count results in a slight deterioration, likely caused by reduced redundancy that normally stabilizes mask prediction.

For $N_p$, our expanded experiments, including extreme settings such as $N_p = 1$ and $N_p = 10$, show that the model exhibits strong robustness to this parameter. While very small $N_p$ may not fully capture object geometry and overly large $N_p$ can distort the prompt distribution, the performance fluctuations among $N_p = 2, 4, 6, 8$ remain minimal. This indicates that the model does not heavily rely on the precise number of point prompts within each group; instead, the prompt embedding remains effective as long as an appropriate amount of spatial cues is preserved.
Overall, $N_q$ serves as the dominant factor influencing instance coverage, whereas $N_p$ introduces only minor variations in performance. Considering both segmentation accuracy and computational efficiency, we adopt moderate settings of $N_q$ and $N_p$ in the final model.

\begin{figure}[!t]
    \centering
    \subfigure[Slight wear]{
    \begin{minipage}[c]{0.28\linewidth}
			\includegraphics[width=1.07in]{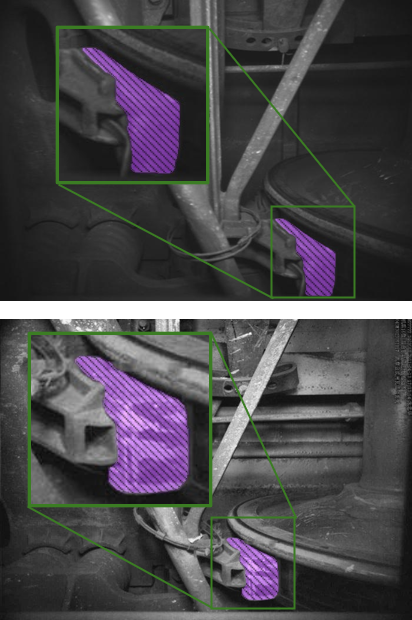}\vspace{0.4em} 
		\end{minipage}
    }
    \subfigure[Moderate wear]{
    \begin{minipage}[c]{0.28\linewidth}
			\includegraphics[width=1.07in]{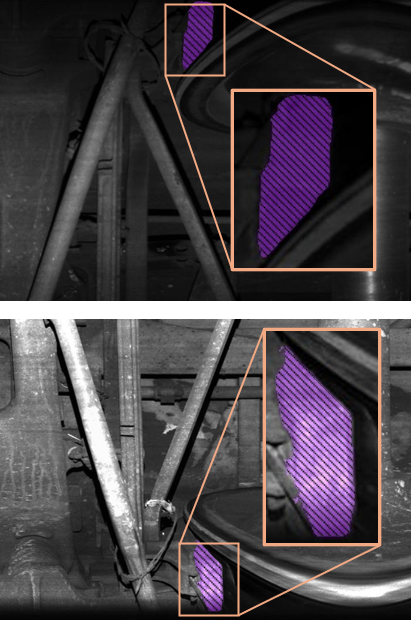}\vspace{0.4em} 
		\end{minipage}
    }
    \subfigure[Severe wear]{
    \begin{minipage}[c]{0.28\linewidth}
			\includegraphics[width=1.07in]{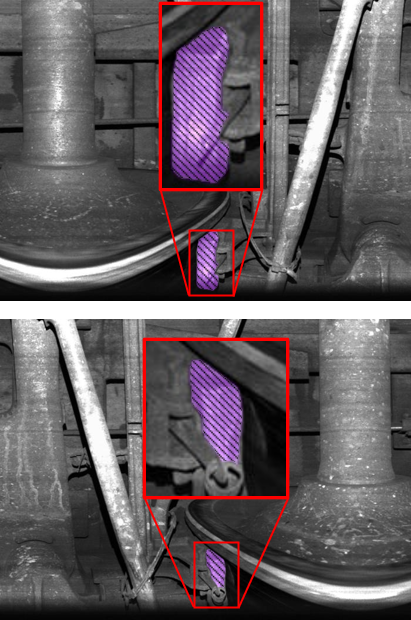}\vspace{0.4em} 
		\end{minipage}
    }
        \caption{Brake shoe wear condition visualization.}
	\label{BST}
\end{figure}
\begin{table}[t]
	\renewcommand{\arraystretch}{1.2}
        \caption{Comparison of detection accuracy across different wear levels using various methods}
        \label{BST_mes}
	\centering
	\footnotesize
	\setlength{\tabcolsep}{1.7mm}{
		\begin{tabular}{c|ccc}
            \toprule
            \textbf{Method} &Slight wear &Moderate wear &Severe wear \\
            \midrule
            SAM FTI-FDet &\textbf{96.7} &\textbf{94.8} &\textbf{97.5}\\
            Mask R-CNN &91.6 &91.2 &93.9\\
            Bounding Box &82.3 &84.1 &88.7\\
            Edge Detection &75.6 &78.4 &81.2\\
            \bottomrule
            \end{tabular}}
\end{table}
\subsubsection{Brake shoe wear measurement}
We employed the SAM FTI-FDet method to evaluate brake shoe wear conditions and compared it with two conventional approaches: one based on projection calculations after obtaining bounding boxes through object detection, and another utilizing geometric analysis following edge detection. The comparative results are presented in Table \ref{BST_mes}, with the detection schematic of SAM FTI-FDet illustrated in Fig. \ref{BST}. The experimental results demonstrate that the instance segmentation–based method significantly outperforms traditional projection and geometric analysis methods in wear detection tasks, with higher-precision segmentation models yielding superior detection performance. This improvement is primarily attributed to the high-quality segmentation results provided by the SAM module integrated within the SAM FTI-FDet framework. For the purpose of evaluation, wear levels are defined as follows: 100\%–70\% is categorized as slight wear, 70\%–30\% as moderate wear, and wear levels below 30\% are considered severe wear, indicating that the brake shoe needs to be replaced in practical operation.

\subsubsection{Impact of Noise and Visual Artifacts}
\begin{table}[t]
	\renewcommand{\arraystretch}{1.2}
        \caption{Quantitative results of different methods with noise and visual artifacts.}
        \label{noise_t}
	\centering
	\footnotesize
	\setlength{\tabcolsep}{0.4mm}{
		\begin{tabular}{c|ccc|ccc}
            \toprule
            \textbf{Method}& $AP^{box}$ &$AP^{box}_{50}$ & $AR^{box}_{50-90}$ &$AP^{mask}$ &$AP^{mask}_{50}$ & $AR^{mask}_{50-90}$ \\
            \midrule
            FastSAM\cite{zhao2023fast}  &56.3 &77.9 &60.5 &57.1 &83.8 &58.3  \\
            RSPrompter\cite{chen2024rsprompter} &55.6 &78.6 &58.8 &55.1 &77.9 &56.4 \\
            Mask R-CNN &51.1 &70.4 &53.6 &50.6 &71.8 &53.8\\
            \textbf{SAM FTI-FDet} &\textbf{60.8} &\textbf{87.6} &\textbf{65.7} &\textbf{59.6} &\textbf{86.6} &\textbf{62.8} \\
            \bottomrule
            \end{tabular}}
\end{table}

\begin{figure}[!t]
    \centering
    \subfigure[]{
    \begin{minipage}[c]{0.28\linewidth}
			\includegraphics[width=1.07in]{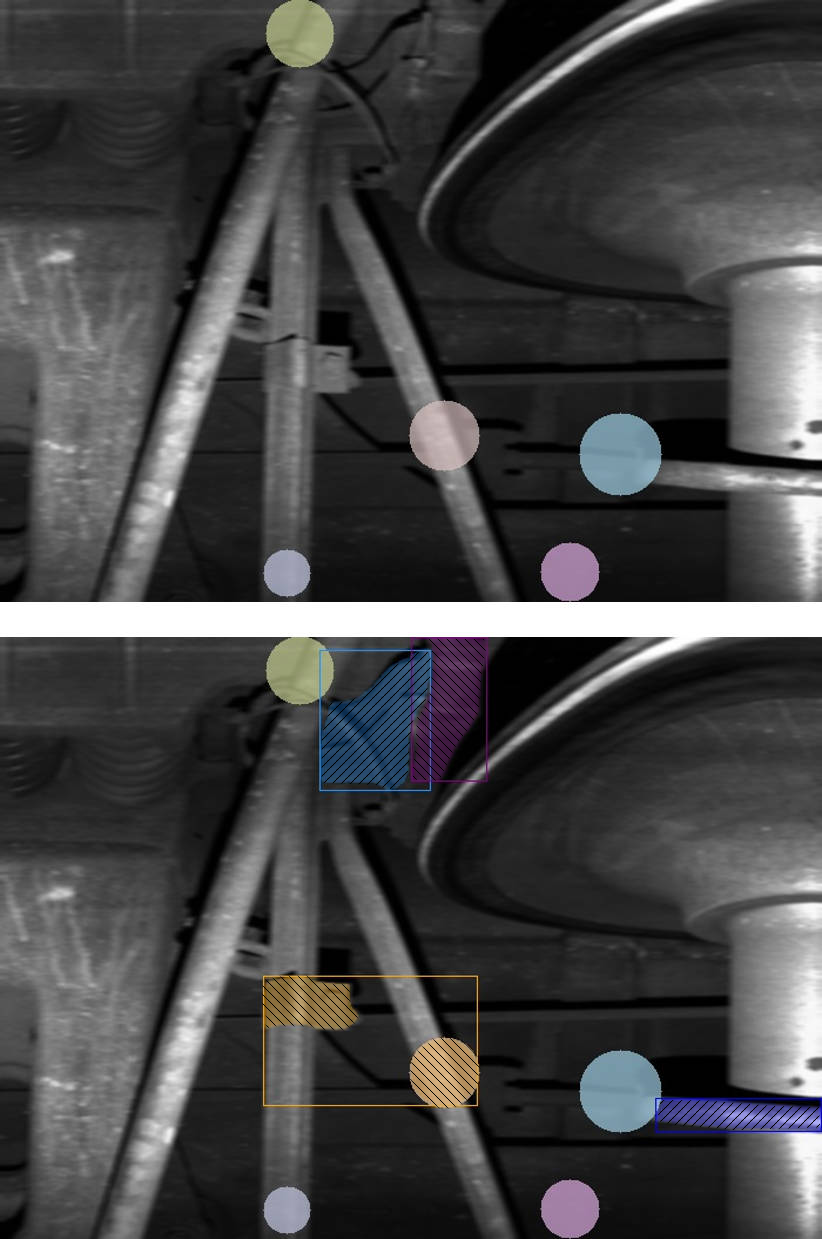}\vspace{0.4em} 
		\end{minipage}
    }
    \subfigure[]{
    \begin{minipage}[c]{0.28\linewidth}
			\includegraphics[width=1.07in]{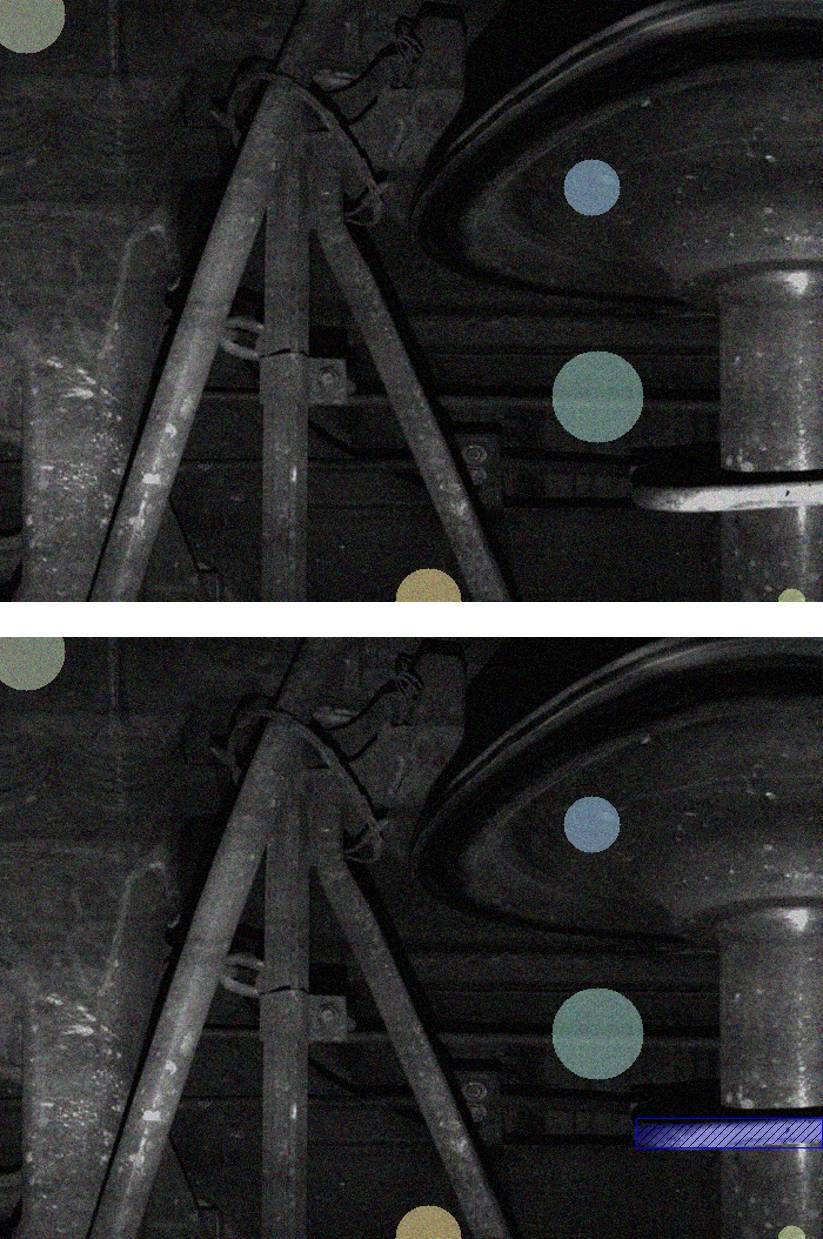}\vspace{0.4em} 
		\end{minipage}
    }
    \subfigure[]{
    \begin{minipage}[c]{0.28\linewidth}
			\includegraphics[width=1.07in]{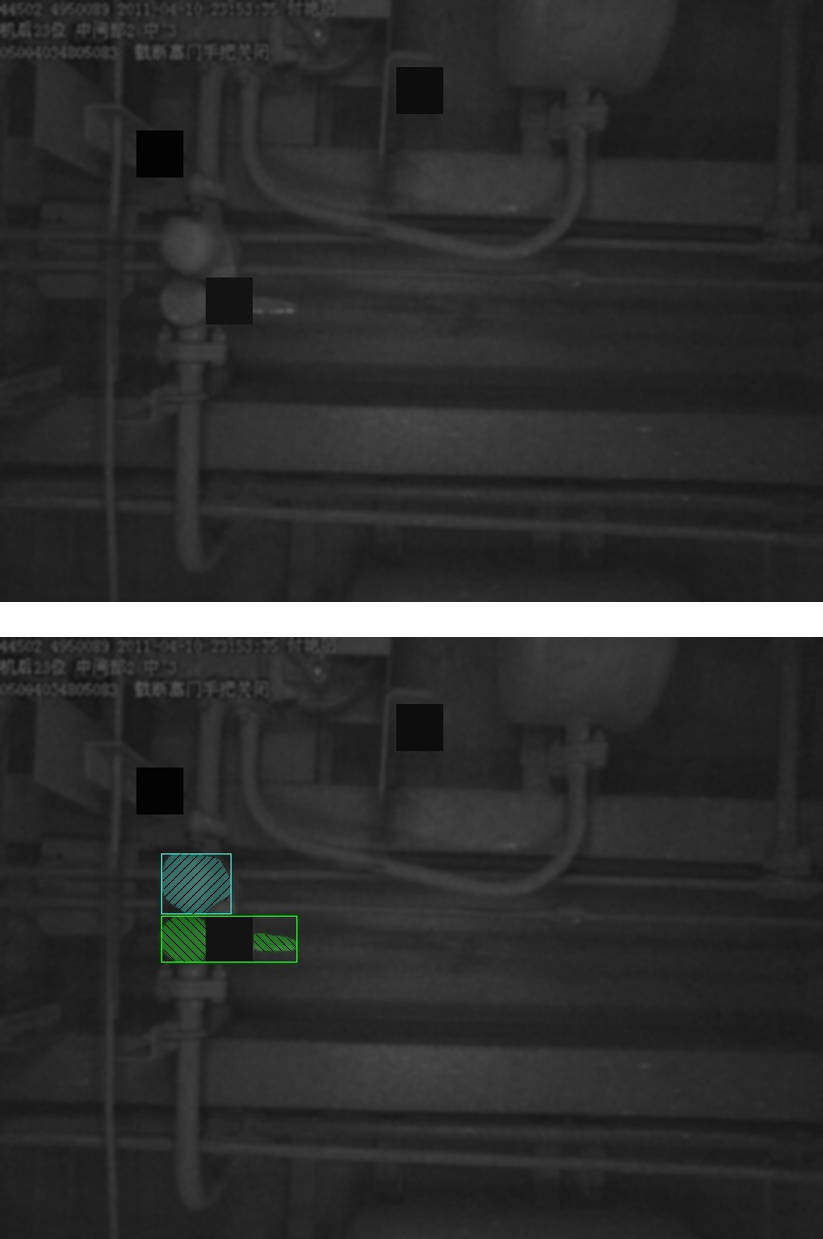}\vspace{0.4em} 
		\end{minipage}
    }
        \caption{Instance detection under various noise and visual artifacts. (a) Motion blur and semi-transparent spots; (b) Gaussian noise and blur; (c) Cutout occlusions and color/contrast variations.}
	\label{noise}
\end{figure}

To evaluate the robustness of our SAM FTI-FDet framework under real-world disturbances, we conducted experiments on the validation set augmented with various noise and visual artifacts, including random semi-transparent spots, cutout-style occlusions, gaussian and salt-and-pepper noise, motion and gaussian blur, as well as contrast and color shifts. The quantitative results show that SAM FTI-FDet achieves the highest performance among all compared methods, with $AP_{box}$ = 60.8, $AP_{mask}$ = 59.6, and corresponding AR values also surpassing the baselines, and these results are detailed in Table \ref{noise_t}.
While other methods such as FastSAM, Rsprompter, and Mask R-CNN suffer severe performance degradation under most challenging conditions, SAM FTI-FDet demonstrates strong robustness against illumination changes and occlusions. However, certain disturbances, particularly Gaussian noise and semi-transparent spots, still lead to noticeable segmentation errors, such as incomplete masks or false positives in background regions, and these typical failure cases are illustrated in Fig. \ref{noise}.
These results demonstrate that the proposed prompt-driven, lightweight SAM framework maintains strong instance-level segmentation accuracy even in the presence of complex visual disturbances, confirming its suitability for robust industrial freight train inspection, while also highlighting potential directions for improving noise-resilient prompt design and mask refinement.

\subsubsection{Generalization on Other Datasets}
\begin{table}[t]
	\renewcommand{\arraystretch}{1.2}
        \caption{Comparisons of different methods on the MS-COCO dataset.}
        \label{coco}
	\centering
	\footnotesize
	\setlength{\tabcolsep}{0.4mm}{
		\begin{tabular}{c|ccc|ccc}
            \toprule
            \textbf{Method}& $AP^{box}$ &$AP^{box}_{50}$ & $AR^{box}_{50-90}$ &$AP^{mask}$ &$AP^{mask}_{50}$ & $AR^{mask}_{50-90}$ \\
            \midrule
            FastSAM\cite{zhao2023fast}  &37.9 &53.6 &50.4 &32.6 &51.8 &49.0  \\
            SAM-seg\cite{chen2024rsprompter} &37.2 &56.1 &47.1 &33.2 &53.3 &42.0\\
            RSPrompter\cite{chen2024rsprompter} &37.6 &54.5 &\textbf{56.3} &35.2&54.3 &\textbf{51.0}\\
            \textbf{SAM FTI-FDet} &\textbf{38.5} &\textbf{57.3} &54.8 &\textbf{36.0} &\textbf{56.6} &49.5 \\
            \bottomrule
            \end{tabular}}
\end{table}
To evaluate the generalization capability of our method, we conducted comparative experiments on the MS-COCO dataset using 8 NVIDIA H100 GPUs, with an input resolution of $512\times512$, an initial learning rate of 0.004, and a batch size of 40. Distinct from the specific model trained on the Freight Train Dataset, the model in this experiment was initialized with pre-trained weights and independently fine-tuned on the MS-COCO dataset. The $512\times512$ input resolution was selected to maintain a scale consistent with the training domain during cross-dataset validation, preserving object sizes, spatial statistics, and visual distributions. This ensures a more objective assessment of the model’s generalization capability. Such a choice also guarantees architectural compatibility and enables fair comparisons within the model family. Although higher resolutions can improve absolute AP values for all methods, they generally do not alter relative performance trends, which is the primary focus of our evaluation in this study.

As shown in Table \ref{coco}, experimental results indicate that the proposed SAM FTI-FDet consistently outperforms existing approaches in both detection and segmentation tasks, achieving a better balance between precision and recall. This performance gain is mainly attributed to the feature-guided prompt generation mechanism, which allocates prompts to structurally salient regions, thereby enhancing instance-specific feature representations.

Although SAM FTI-FDet exhibits superior AP and precision, its slightly lower AR can be explained by the characteristics of the prompt allocation strategy. Qualitative analysis of the predictions shows that missed objects are typically small, low-contrast, or have blurred boundaries—features that are inherently less visually salient. With a fixed number of prompts, such objects are more likely to receive insufficient prompt coverage. In contrast, highly salient categories, such as people, cars, and dogs, consistently receive prompts and are accurately detected and segmented. This phenomenon further supports the notion that limited prompt resources naturally prioritize the more prominent structural regions in an image.

\section{Discussion}
The proposed SAM FTI-FDet framework demonstrates significant improvements in instance-level fault detection for freight train components, primarily driven by the integration of a query-based prompt generator and a lightweight TinyViT-SAM backbone. Unlike traditional box-based prompts that often introduce background noise, our query-based mechanism enables precise semantic guidance for the mask decoder, enhancing the alignment between image features and the semantic information of target components. This mechanism effectively mitigates the limitations of conventional methods, allowing the model to accurately delineate boundaries even in scenarios characterized by complex structural layouts, frequent occlusions, and background clutter.

A critical finding of this study is the framework's strong generalization capability across vastly different data scales, as evidenced by results on both the small-scale Freight Train Dataset and the large-scale MS-COCO dataset. Rather than implying a single model fits all, this generalization reflects the high architectural adaptability of our framework under independent fine-tuning. On the small-scale Freight Train Dataset, the strategy of freezing the pre-trained Foundation Model backbone acts as a strong regularizer, effectively preventing the lightweight decoder from overfitting to limited samples. Conversely, on the large-scale MS-COCO dataset, the proposed prompt-driven mechanism proves to possess sufficient representational capacity to capture complex, general object features. This dual-effectiveness confirms that our architecture strikes a robust balance: compact enough to be stable on specialized industrial tasks, yet expressive enough to scale to diverse general scenarios.

Beyond accuracy and generalization, the framework is optimized for practical deployment efficiency. The integration of the TinyViT-SAM backbone significantly reduces computational overhead and memory consumption, making real-time processing feasible on edge-side monitoring devices. To further enhance robustness in visually challenging conditions, the Adaptive Feature Dispatcher combines global feature aggregation with local feature decomposition. This design ensures that the model retains multi-scale representations of components, maintaining segmentation stability even in the presence of environmental interferences such as dust or uneven illumination.

From an industrial perspective, SAM FTI-FDet offers a deployable and interpretable solution for predictive maintenance. By providing fine-grained, instance-level segmentation, the method extends beyond simple fault identification to support quantitative assessment, such as measuring brake shoe thickness. This capability enables data-driven maintenance decisions, reducing manual inspection costs. Moreover, the framework’s adaptability suggests potential for broader application across various industrial inspection tasks, including other railway equipment or machinery with repetitive structural patterns.

Despite these advantages, certain limitations remain. The query-based prompt mechanism may still struggle with extremely small or low-saliency defects, potentially leading to missed detections. Additionally, severe occlusions or highly cluttered backgrounds can impact segmentation quality, highlighting the need for further refinement of prompt allocation strategies. Although the model is lightweight, the training of the prompt generator introduces additional complexity. Future work will focus on enhancing real-time multi-fault detection, optimizing prompt generation for low-saliency targets, and exploring multi-modal data integration to further improve system robustness and scalability.

\section{Conclusion}
This paper presents SAM FTI-FDet, a prompt-driven instance segmentation framework that adapts the Segment Anything Model to automated freight train fault detection. By introducing an end-to-end query-based prompting mechanism and a lightweight TinyViT-SAM backbone, the proposed framework enables SAM to autonomously acquire task-relevant spatial and semantic cues without relying on manually designed category-specific prompts. This design effectively bridges the gap between large-scale foundation model pretraining and domain-specific industrial inspection requirements, while maintaining a compact model size suitable for edge-side deployment.
Extensive experiments on real-world freight train datasets demonstrate that SAM FTI-FDet achieves superior instance-level segmentation accuracy and strong generalization capability under transfer learning, outperforming existing SAM-based and conventional segmentation methods with significantly reduced computational overhead. The results indicate that efficient prompt learning provides an effective pathway for injecting domain knowledge into foundation models, enabling robust adaptation to complex industrial environments characterized by structural repetition, occlusion, and background clutter.
Looking forward, future work will focus on extending the proposed framework to temporal modeling for video-based fault detection and real-time deployment on embedded inspection platforms, aiming to further enhance its scalability and applicability in practical railway monitoring systems. More broadly, this study highlights the potential of prompt-driven foundation model adaptation as a general paradigm for industrial visual inspection tasks beyond the railway domain.

\bibliographystyle{IEEEtran}
\bibliography{reference}\ 

\vfill

\end{document}